\documentclass{article}

\usepackage[nonatbib, preprint]{neurips_2024}

\usepackage[utf8]{inputenc} 
\usepackage[T1]{fontenc}    
\usepackage{hyperref}       
\usepackage{url}            
\usepackage{booktabs}       
\usepackage{amsfonts}       
\usepackage{nicefrac}       
\usepackage{microtype}      
\usepackage{xcolor}         
\usepackage{amsmath}
\usepackage{mathrsfs}
\usepackage{graphicx}
\usepackage{pifont}
\usepackage{wrapfig}
\usepackage{ifsym}
\usepackage[ruled,linesnumbered]{algorithm2e}
\SetKwComment{Comment}{/* }{ */}
\usepackage{todonotes}

\usepackage{booktabs}
\usepackage{units}

\usepackage{subfigure}

\usepackage{graphicx}
\usepackage{amsmath}
\usepackage{amssymb}
\usepackage{booktabs}
\usepackage{amsmath, bm, epstopdf, url, pifont, overpic, cases}
\usepackage{url}
\usepackage{booktabs}
\usepackage{amssymb}
\usepackage{bbding}
\usepackage {pifont}
\usepackage{wasysym}
\usepackage{utfsym}
\usepackage{fontawesome}
\usepackage{multirow}
\usepackage{bbding}
\usepackage{pifont}
\usepackage{wasysym}

\usepackage{colortbl}
\usepackage{xcolor}
\usepackage{color}
\usepackage{array}  
\usepackage{enumitem}
\usepackage{xfrac} 
\usepackage{makecell}
\usepackage{url}

\newcommand{\boldblue}[1]{\textcolor{blue}{\textbf{#1}}}

\newcolumntype{C}{@{\hspace{0.1em}}c@{\hspace{0.1em}}}
\newcolumntype{D}{@{\hspace{0.2em}}c@{\hspace{0.2em}}}
\newcolumntype{T}{@{\hspace{0em}}c@{\hspace{0em}}}

\setlist[itemize]{leftmargin=*}
\definecolor{color3}{rgb}{0.95,0.95,0.95}

\title{Protect-Your-IP: Scalable Source-Tracing and Attribution against Personalized Generation}

\author{
 Runyi Li\textsuperscript{1}, 
 Xuanyu Zhang\textsuperscript{1}, 
 Zhipei Xu\textsuperscript{1}, 
 Yongbing Zhang\textsuperscript{3}, 
 Jian Zhang\textsuperscript{1,2 \Envelope} \\
 \textsuperscript{1} School of Electronic and Computer Engineering, Peking University\\
 \textsuperscript{2} Peking University Shenzhen Graduate School-Rabbitpre AIGC Joint Research Laboratory\\
 \textsuperscript{3} School of Computer Science and Technology, Harbin Institute of Technology (Shenzhen)\\ 
}

\begin{document}

\maketitle
\renewcommand*{\thefootnote}{\Envelope}
\footnotetext[1]{Corresponding author, zhangjian.sz@pku.edu.cn.}

\begin{abstract}
With the advent of personalized generation models, users can more readily create images resembling existing content, heightening the risk of violating portrait rights and intellectual property (IP). Traditional post-hoc detection and source-tracing methods for AI-generated content (AIGC) employ proactive watermark approaches; however, these are less effective against personalized generation models. Moreover, attribution techniques for AIGC rely on passive detection but often struggle to differentiate AIGC from authentic images, presenting a substantial challenge.
Integrating these two processes into a cohesive framework not only meets the practical demands for protection and forensics but also improves the effectiveness of attribution tasks. Inspired by this insight, we propose a unified approach for image copyright source-tracing and attribution, introducing an innovative watermarking-attribution method that blends proactive and passive strategies. We embed copyright watermarks into protected images and train a watermark decoder to retrieve copyright information from the outputs of personalized models, using this watermark as an initial step for confirming if an image is AIGC-generated.
To pinpoint specific generation techniques, we utilize powerful visual backbone networks for classification. Additionally, we implement an incremental learning strategy to adeptly attribute new personalized models without losing prior knowledge, thereby enhancing the model’s adaptability to novel generation methods.
We have conducted experiments using various celebrity portrait series sourced online, and the results affirm the efficacy of our method in source-tracing and attribution tasks, as well as its robustness against knowledge forgetting.
\end{abstract}

\section{Introduction}
\label{sec:introduction}
The AI-generated content (AIGC) model, especially the personalized generation model~\cite{zhang2024surveypersonalized}, has adverse implications on the copyright and intellectual property (IP) of various visual content such as artworks by artists, portraits of individuals, and photographs by photographers. Images generated through personalized methods may propagate misinformation and infringe upon copyrights, thereby engendering negative societal repercussions.
In response to these concerns, significant research efforts have been directed towards AIGC copyright watermarking for images~\cite{van2023anti,liang2023mist,zhang2024v2a}. Several methodologies, including those employing box-free watermarking techniques~\cite{yu2020responsible,wu2020watermarking,9373945,huang2023can,tan2023deep,wu2023sepmark,ma2023unified,wang2024must}, have been developed to trace back the fine-tuning of generative models like Generative Adversarial Network (GAN)~\cite{goodfellow2020generative,karras2017progressive,gulrajani2017improved,brock2018large} and Diffusion models~\cite{ho2020denoising, song2021denoising,rombach2021highresolution,dhariwal2021diffusion,song2023consistency}, thereby serving a post-hoc protective function.

For AIGC models, particularly personalized generation models, both the \textbf{source-tracing} of copyright and the \textbf{attribution} of specific methods are important, and need to be accomplished simultaneously~\cite{wang2023security}. Given the inherent difficulty in preventing the personalized generation of portraits at the source, we opt for a post-hoc evidentiary approach. This approach not only supports copyright source-tracing of generated outputs but also enables the determination of whether an image is AIGC-generated and attribution of the specific personalized generation method employed by the stealer, thus facilitating attribution tasks.
\begin{figure}
    \centering
    \includegraphics[width=1\linewidth]{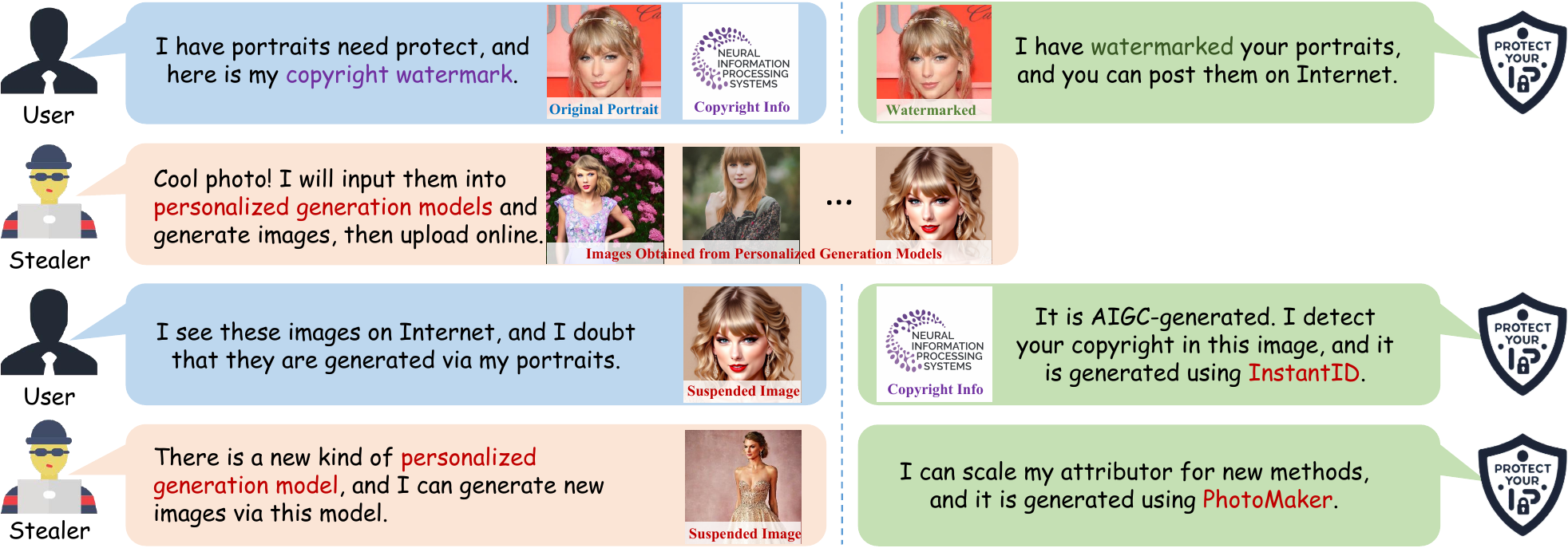}
    \caption{Our proposed protecting scenario, wherein users embed invisible watermarks to their images in advance. If suffering stealing via personalized generative methods, users can defend their rights via the copyright information and attributed generation method provided by us.}
    \label{fig:teaser}
    \vspace{-6mm}
\end{figure}

However, existing research struggles to simultaneously accomplish the aforementioned tasks of source-tracing and attribution. 
Proactive watermark methods~\cite{9373945, ma2023unified} involve embedding invisible watermarks into generated outputs to identify copyright and generation method information. However, this approach is incongruent with our scenario, as stealers are unlikely to add their own watermarks to personalized generation models voluntarily. 
Passive detection methods~\cite{li2024regeneration,xu2024detecting} align with our threat model; however, their performance suffers in distinguishing between AIGC-generated and real images and in attributing different personalized generation methods. Furthermore, considering the continual emergence of newly appeared methods, we also require a scalable approach capable of incrementally updating existing models to address newly proposed personalized generation methods.

To clarify our task and scenario, we clarify the definition of source-tracing and attribution as illustrated in Fig.~\ref{fig:teaser}: (1) \textbf{Source-Tracing}: "Who does this suspended image belong to?" We aim to accurately retrieve the copyright information of an image; (2) \textbf{Attribution}: "Is this generated by personalized model? If so, by which model?" We aim to first determine whether it is generated via the decoded result, and then judge the specific generation method. 

Based on the aforementioned challenges, we propose a novel framework for source-tracing and attribution of personalized generated images, employing a combination of \textbf{proactive} watermark and \textbf{passive} detection mechanisms. We embed portrait images requiring protection with box-free watermarks. Specifically tailored for personalized generation methods, our copyright watermark can be decoded from the results of personalized generation, allowing for the determination of whether a suspicious image is AIGC-generated based on the presence or absence of the watermark. This constitutes the proactive source-tracing watermark, thereby achieving initial attribution for AIGC content detection. For attribution of specific personalized generation methods, we employ a classification network. Furthermore, recognizing the continuous evolution of personalized generation methods, we utilize incremental learning techniques, enabling existing models to attribute new generation methods with minimal data and training costs. Fig.~\ref{fig:teaser} shows a scenario of our protecting framework. Our main contributions are summarized as follows:

\vspace{2pt}
\noindent \ding{113}~(1) We propose the task of source-tracing and attribution of personalized generation, and design a unified framework that combines proactive watermarking and scalable passive detection.

\vspace{2pt}
\noindent \ding{113}~(2) For the source-tracing task, we apply an encoder-decoder structure to embed invisible source-tracing watermarks into images, which can be decoded from the generated results. Through the decoded result of the suspended image, we can identify the copyright, and determine whether the image is personalized generated.

\vspace{2pt}
\noindent \ding{113}~(3) For the attribution task, we introduce a hierarchical attribution approach. Leveraging the proactive watermark, we first determine the presence of a copyright watermark, and then employ a visual backbone to figure out the specific generation method. Furthermore, we incorporate incremental learning strategy into this network, enabling scalable attribution of new personalized methods.

\vspace{2pt}
\noindent \ding{113}~(4) Experimental results demonstrate the effectiveness of our proposed methods in source-tracing and attributing tasks.


\section{Related work}
\label{sec:related work}

\subsection{Model Watermarking Methods}
Several varieties of watermarking techniques have been proposed to ascertain the ownership of models and embed owner information within them. These include \textit{white-box} methods that add watermarks to model weights or model outputs\cite{uchida2017embedding,chen2019deepmarks}. Assuming the model protector is familiar with the model architecture, white-box watermarks can be embedded into the model's weights, and the decoded watermark can then represent the identity of the owner. Conversely, if the owner does not understand the model structure, a \textit{black-box} watermark\cite{adi2018turning} can be added to the model's output, such as by constructing a trigger set, to indicate ownership.
Furthermore, in some special scenarios where there is no predefined model structure or the scenario is not related to specific model structure, this is referred to as \textit{box-free} watermarking\cite{yu2020responsible,wu2020watermarking,9373945,huang2023can,tan2023deep,wu2023sepmark}. The method proposed by Zhang et al.~\cite{9373945}. involves adding specific watermarks to the image processing models. Specifically, we can first add a watermark to the image. The image with the watermark, after being processed by the model, yields an output result from which the watermark can be decoded. Additionally, for images that have not undergone processing by this model, a benign signal will be decoded instead. Our proposed method applies this box-free design, as detailed in Sec.~\ref{sec:method_box_free}.
\subsection{Attribution of Generative Models}
To construct a complete chain of evidence in digital forensics, it is not only essential to understand the copyright of an image but also to be interested in the specific generative methods used to produce pirated images. Given a suspicious image, the inference of whether the image is generated by AIGC and the identification of the specific AIGC method that produced the result is referred to as attribution. Current passive judgment regarding AIGC images primarily involves designing special network structures to perform a binary classification between real and fake images, or predict a suspicious map indicating generation trace~\cite{dutta2021ensembledet, ge2022explaining, coccomini2022combining}.
For the judgment of specific generative methods, it mainly includes two approaches: classification~\cite{goebel2020detection,yang2022deepfake,girish2021towards} and finger-printing~\cite{yu2018attributing,yu2021artificial, yang2021learning, yu2020responsible,pmlr-v202-nie23a}. The classification method generally uses a universal visual network as its backbone, combined with special designs, to achieve differentiation among various generative methods. Finger-printing, on the other hand, focuses on the different characteristics of the results generated by different methods, analyzing the results in terms of frequency domains, feature spaces, and other aspects.
To our best knowledge, there have been a few explorations of the finger-printing method regarding the attribution of diffusion-based generation models~\cite{kim2023wouaf,nie2023attributing,wen2024tree}, but these approaches add watermarks to generative models \textit{in advance}, and thus do not align with our scenario.


\section{Problem Statement \& Threat Model}
The task of intellectual property (IP) protection proposed by us encompasses two main components: (1) source-tracing the copyright of results generated by personalized generation methods to determine the ownership of the image, and (2) distinguishing between suspicious images found online as legal real photographs or illegal results obtained through personalized generation methods. In the case of illegal results, additional identification of the specific model used for generation (such as LoRA~\cite{hu2022lora}, InstantID~\cite{wang2024instantid}, etc.) is required.

\ding{113} \textbf{Task and Method Definition}
For the \textit{source-tracing} task, we aim to identify the copyright of suspended images from the Internet. For the \textit{attribution} task, we aim to first determine whether the image is AIGC-generated, and if so, we further judge the specific generation method of the image. For \textit{box-free} watermarking, it means a kind of watermarking technique that embeds invisible watermarks into the model 
outputs, and both the model and the outputs can be protected.

\ding{113} \textbf{Stealer's Objective} The stealer pertains to the acquisition of a collection of images requiring copyright protection, such as celebrity portraits, animated characters, and endeavors to utilize personalized generation models to produce images containing features resembling those present in the aforementioned images, such as facial characteristics.

\ding{113} \textbf{Stealer's Knowledge} 
The stealer's objective is limited to images and does not involve models or network architectures. Therefore, the scenario we set up is a "box-free" environment, wherein both the watermark we add and the process of the stealer's attack do not involve the structure of the model. Furthermore, the stealer is unaware of the method we use to add the watermark. Given that the current situation involves image IP theft by the stealer, proactive model watermarking is not applicable to our scenario (it is obvious that stealers would not voluntarily add watermarks to their own models).

\ding{113} \textbf{Stealer's Capability} The stealer has access to the Internet and can obtain the images we need to protect, as well as ample resources to train personalized generation models. 

\ding{113} \textbf{Security Requirements} Our proposed watermark should satisfy two requirements: (1) robustness against potential degradation during transmission over networks, such as noise, JPEG compression, and other forms of degradation; (2) preservation of the quality of the image, ensuring that the watermark itself does not impact the visual effect of the image.
\section{Methods}
\label{sec:methods}

\subsection{Overview}
\paragraph{Motivation}
Current proactive forensic watermarking primarily targets non-personalized image generation models, which lacks research pertaining to forensic evidence of results generated by personalized generation models.
Existing post-hoc methods for image IP protection mainly rely on proactive forensic watermarking techniques. The efficacy of proactive forensic watermarking typically entails detecting copyright watermarks from suspicious images, while it does not provide insight into which specific AIGC model generated the image, thereby hindering the establishment of comprehensive forensic evidence in practice. 

For attribution tasks involving determining whether an image is AIGC-generated and inferring the specific generation method, current approaches mainly rely on passive detection methods such as classification~\cite{xu2024detecting,goebel2020detection} and fingerprinting~\cite{yu2021artificial}. However, existing AIGC models can generate highly realistic images, making it challenging to discern their authenticity solely through passive approaches. 

\begin{figure}
    \centering
    \includegraphics[width=1\linewidth]{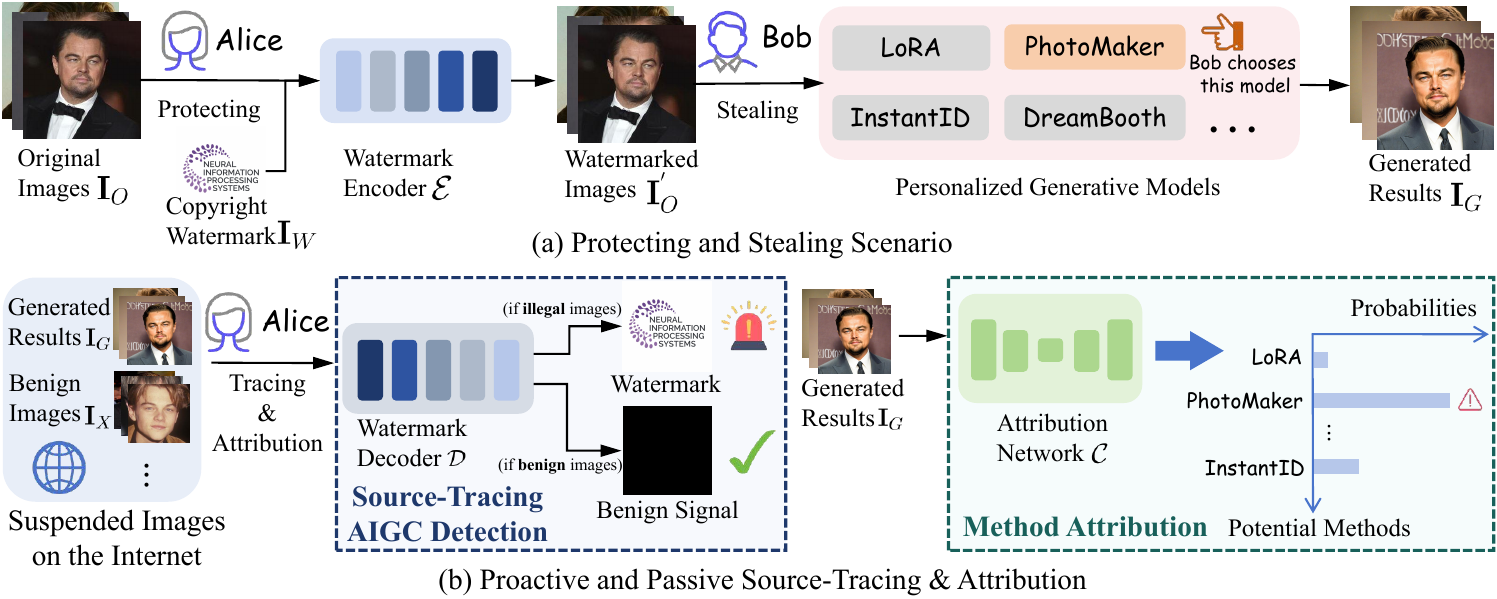}
    \vspace{-2mm}
    \caption{Overview of our proposed watermarking framework for image IP protection. (a) shows the process of IP owner Alice adding invisible copyright watermark to images, and stealer Bob training personalized generative models using watermarked images. (b) shows Alice dealing with suspended images from Internet. First Alice trying to source-trace the copyright of the image, and if the watermark decoder outputs a benign signal, it is not generated via personalized generative models, otherwise it will output the copyright watermark. Alice further judges by which model it is generated.}
    \label{fig:overview}
\end{figure}

Motivated by the aforementioned insights, we propose a unified task of copyright source-tracing and image attribution and employ a combined approach utilizing both proactive watermarking and passive detection methods. By unifying these two tasks, we meet the need of realistic protection forensics, and allows for better performance of the attribution task.
\vspace{-3mm}
\paragraph{Pipeline}
The user adds a watermark image $\mathbf{I}_W$ to the original images $\mathbf{I}_O$, resulting in the watermarked image denoted as $\mathbf{I}_O^{'}$. This watermarked image can be publicly uploaded on the Internet. A stealer obtains images from $\mathbf{I}_O^{'}$ from the Internet and illegally generates additional images using a personalized generation model, denoted as $\mathbf{I}_G$. There are also legal images on the Internet similar to $\mathbf{I}_O^{'}$, denoted as $\mathbf{I}_X$. When the users encounter suspicious images on the Internet, they can decode them using a watermark decoder $\mathcal{D}$. Personalized generated images can be decoded to reveal the copyright watermark, while images from $\mathbf{I}_O$ and $\mathbf{I}_X$ decode to a black image, serving as a benign signal. If the copyright watermark is decoded, the user can further utilize our method to determine which personalized generation model generated the image via attribution network $\mathcal{C}$, thus completing the evidential chain. An illustration of our proposed framework is presented in Fig.~\ref{fig:overview}.

\subsection{Proactive Watermark for Source-Tracing}
\paragraph{Model Watermarking Methods}
\label{sec:method_box_free}
To safeguard copyright and IP in images, we construct a model-agnostic protection scenario and implement watermarking without box embedding for copyright information embedding and source-tracing. Specifically, we embed the watermark information into the output distribution of the image processing model through a network and extract the watermark using a corresponding extraction network. Besides considering robustness against traditional digital image processing, our approach also defends against model extraction attacks, as box-free watermarking does not require the extractor to have knowledge of the target model's internal details.
\paragraph{Training and Inference of Watermark}
Given a set of original images requiring protection, denoted as $\mathbf{I}_O$, we apply imperceptible image watermarks using an embedding network, resulting in the watermarked images denoted as $\mathbf{I}_O^{'}$. The watermark encoder $\mathcal{E}$ is implemented using a reversible network~\cite{zhang2023editguard}, and the details of network architecture and training process, including loss function, are in Appendix~\ref{sec:network_architecture}. These watermarked images are then publicly uploaded online. Subsequently, stealers download these images and utilize them to train personalized generation models such as LoRA~\cite{hu2022lora}. The generated results, denoted as $\mathbf{I}_G$, can be decoded using decoding network $\mathcal{D}$ to obtain the embedded image watermark, enabling source-tracing of copyright and IP. The watermark decoding network shares the same structure as the aforementioned watermark embedding network. Due to the reversible design of the network, the decoding process is the inverse of the embedding process. Additionally, when decoding legal images $\mathbf{I}_X$, the decoding network should output a black image $\mathbf{0}$, serving as a benign signal.
The loss function could be expressed as Eq.~\ref{eq:source_loss}:
\begin{equation}
    \label{eq:source_loss}
    \mathcal{L}_{\text{decoder}} = \sum_{i=1}^{n}||\mathcal{D}(\mathbf{I}_{G_{i}})-\mathbf{I}_{W}||_2^2
    + ||\mathcal{D}(\mathbf{I}_O)-\mathbf{0}||_2^2
    + ||\mathcal{D}(\mathbf{I}_{X})- \mathbf{0}||_2^2
\end{equation}
where $\mathbf{I}_{G_{i}}$ denotes images generated by $i^{\text{th}}$ generation method $\mathcal{G}_{i}$, and there are $n$ generation methods adapted in training. A detailed training process is shown in Algo.~\ref{alg:pipeline}.
\begin{figure}[t]
\vspace{-3mm}
\begin{minipage}[t]{.50\textwidth}
\begin{algorithm}[H]
    \scriptsize
    \caption{\footnotesize{Training of Watermark Decoder}}
    \label{alg:pipeline}
    \KwIn{$\mathbf{I}_{O}$, $\mathbf{I}_{W}$, $\mathbf{I}_{X}$, $\mathbf{0}$, $\mathcal{E}$, $\mathcal{D}$, $\mathcal{G}_{1,2,...,n}$}
    \KwOut{Watermark Decoder $\mathcal{D}$}
    $\mathbf{I}_{O}^{'} = \mathcal{E}(\mathbf{I}_{O})$ \\
    \For{$i=1$ \KwTo $n$}{
        $\mathbf{I}_{G_{i}} = \mathcal{G}_{i}(\mathbf{I}_{O}^{'})$ 
    }
    \For{epoch in epochs}{
    Calculate loss $\mathcal{L}_{\text{decoder}}$ using Eq.~\ref{eq:source_loss} \\
    \label{line:original_loss}
        $\mathcal{L}_{\text{decoder}}\text{.backward()}$\\
        $\mathcal{D}\text{.update()}$
    }
    \Return $\mathcal{D}$
    \end{algorithm}
    \vspace{-2mm}
\end{minipage}
\begin{minipage}[t]{.49\textwidth}
\begin{algorithm}[H]
    \scriptsize
    \caption{\footnotesize{Incremental Fine-tuning Strategy for Scalable Attribution}}
    \label{alg:increment}
    \KwIn{$\mathbf{I}_{O}^{'}$, $\mathbf{G}=\{ \mathbf{I}_{G_{1}},\mathbf{I}_{G_{2}},...,\mathbf{I}_{G_{n+1}} \}$, $\mathcal{G}_{1,2,...,n}$, $\mathcal{C}$}
    \KwOut{Fine-tuned Attribution Network $\mathcal{C}$}
    \For{$i=1$ \KwTo $n$}{
        $\mathbf{I}_{G_{i*}} = \mathcal{G}_{i}(\mathbf{I}_{O}^{'})$ 
    }
    \For{epoch in epochs}{
        Calculate loss $\mathcal{L}_{\text{increment}}$ using Eq.~\ref{eq:increment}. \\
        $\mathcal{L}_{\text{increment}}\text{.backward()}$\\
        $\mathcal{C}\text{.update()}$
    }
    \Return $\mathcal{C}$
    \end{algorithm}
\end{minipage}
\vspace{-3mm}
\end{figure}
To simulate the potential degradations that may occur during image transmission in real-world network environments, we have introduced three types of degradations to our training data, which include Gaussian noise, Poisson noise, and JPEG compression. The specific settings for these degradations are detailed in Sec.~\ref{sec:implementation_details}.


\subsection{Scalable Proactive and Passive Attribution}
\paragraph{Hierarchical Proactive-and-Passive Mechanism for Attribution}
\label{sec:attribution}
Our attribution task consists of two components: (1) determining whether suspicious images found on the Internet are generated by personalized generation models, and (2) if indeed generated by an AIGC model, identifying the specific model used. Since we can detect watermarks on AIGC-generated results, we can design a mechanism that combines proactive and passive methods. Initially, we utilize proactive watermarks, which are relatively easy to detect, to differentiate between benign legal images and AIGC-generated results. Subsequently, we employ passive methods to attribute the specific generation method.

For discerning the specific generation method, we draw inspiration from~\cite{xu2024detecting} and utilize an efficient visual backbone to classify the results generated by different methods. The loss function is as Eq.~\ref{eq:loss_att}:
\begin{equation}
    \label{eq:loss_att}
    \mathcal{L}_{\text{attribution}} = \sum_{i=1}^{n}\ell (\mathcal{C}(\mathbf{I}_{G_{i}}), G_{i})
\end{equation}
where $\ell$ is cross-entropy loss, $\mathcal{C}(\cdot)$ is the attribution classifier, and $G_{i}$ is the label of image $\mathbf{I}_{G_{i}}$.
\paragraph{Increment Learning Strategy for Scalable Attribution}
\label{sec:exp_increment}
In real-world scenarios, the types of personalized generation models are constantly evolving and updating. If we choose to fine-tune the existing model every time we encounter a new personalized generative model, this approach will result in a significant training cost and will encounter the catastrophic forgetting problem brought by training on new tasks. To efficiently attribute new generation methods without suffering from catastrophic forgetting problem, we can employ the regularization-based incremental learning strategy~\cite{wang2024comprehensive}, wherein we preserve the knowledge of the original model while learning knowledge of the new generation methods. Following the regularization method of generating new training samples~\cite{li2017learning}, we apply the following training strategy:

(1) Prepare an extra dataset $\mathbf{G}_* = \{\mathbf{I}_{G_{1*}}, \mathbf{I}_{G_{2*}},...,\mathbf{I}_{G_{n*}}\}$ by generating new images using $\mathbf{I}_O^{'}$, where $\mathbf{I}_{G_{i*}}$ is generated by $\mathcal{G}_{i}$, and the scale of $\mathbf{G}_*$ is smaller than the training dataset $\mathbf{G} = \{\mathbf{I}_{G_{1}}, \mathbf{I}_{G_{2}},...,\mathbf{I}_{G_{n+1}}\}$ (including images generated by new appeared personalized model). In our experiments, the length of $\mathbf{G}_*$ is $\frac{1}{5}$ of dataset $\mathbf{G}$. 

(2) Fine-tuning the attribution network using $\mathbf{G}_*$ and $\mathbf{G}$, via the following loss function as Eq.~\ref{eq:increment}:
\begin{equation}
    \label{eq:increment}
    \mathcal{L}_{\text{increment}} =  \lambda_c \; \sum_{i=1}^{n}  \ell(\mathcal{C}(\mathbf{I}_{G_{i*}}), G_{i*}) +  \sum_{i=1}^{n+1} \ell (\mathcal{C}(\mathbf{I}_{G_{i}}), G_{i})
\end{equation}
Note that there are $(n+1)$ kinds of generation methods now. We set $\lambda_c = 1.0$ as the weight of the incremental part loss, and the related ablation studies are shown in Sec.~\ref{sec:lambda_c}. Through the way of adding a light-weight extra dataset into training process, the original attribution is able to learn new knowledge without suffering from forgetting previous task. The detailed process is shown in Algo.~\ref{alg:increment}.


\section{Experiments}
\label{sec:experiment}

\subsection{Implementation Details}
\label{sec:implementation_details}

\begin{table} 
\caption{Evaluation of source-tracing watermark. $\text{PSNR}_\text{embed}$ denotes the Peak Signal-to-Noise Ratio (PSNR) of the watermarked image and the original image, and $\text{PSNR}_\text{decode}$ denotes the PSNR of decoded image from generated results and target watermark. "Zhang et al.~\cite{9373945}$^{\dagger}$" indicates the box-free approach in Zhang et al.~\cite{9373945}, with our watermark encoder and decoder architecture. Best results are shown in \textbf{\textcolor{blue}{bold}}.}
\tiny
\label{tab:main}
\centering
\resizebox{0.75\linewidth}{!}{
\begin{tabular}{cc|cc|cc} 
\toprule
\multirow{2}{*}{Portrait IP} & \multirow{2}{*}{Method} & \multicolumn{2}{c|}{Watermark: \textit{NeurIPS}}  & \multicolumn{2}{c}{Watermark: \textit{IceShore}} \\
& &$\text{PSNR}_\text{embed}$ & $\text{PSNR}_\text{decode}$ &$\text{PSNR}_\text{embed}$ &$\text{PSNR}_\text{decode}$ \\
\midrule
\multirow{3}{*}{Diva} 
&Zhang et al.~\cite{9373945}& 42.4496 & 14.7672  & 46.4082 & 17.1526  \\
&Zhang et al.~\cite{9373945}$^{\dagger}$& 43.7054 & 28.2597  & 47.4658 & 38.4593  \\
&Ours  & \boldblue{49.3972}  & \boldblue{38.2023}  & \boldblue{50.7175} & \boldblue{48.9043} \\
\midrule
\multirow{3}{*}{Sportsman} 
&Zhang et al.~\cite{9373945} &  36.4110 &  14.7998 & 39.6772 & 16.8941  \\
&Zhang et al.~\cite{9373945}$^{\dagger}$& 41.5621 & 28.3615  & 40.7506 & 37.8944  \\
&Ours&\boldblue{50.2480}   & \boldblue{44.9456}  & \boldblue{50.8997} & \boldblue{44.2858} \\
\midrule
\multirow{3}{*}{Actor}   
& Zhang et al.~\cite{9373945}  & 41.0348 & 16.7669 & 44.5886 & 12.4662 \\
&Zhang et al.~\cite{9373945}$^{\dagger}$& 42.9745 & 28.3445  & 45.8301 & 37.9649  \\
& Ours  & \boldblue{50.2719} & \boldblue{53.7781}  & \boldblue{50.4068} & \boldblue{47.7241} \\
\midrule
\multirow{3}{*}{Actress}  
& Zhang et al.~\cite{9373945} &37.3238 & 14.8029 & 35.0916 & 17.0501  \\
&Zhang et al.~\cite{9373945}$^{\dagger}$& 39.3804 & 28.0742  & 37.2112 & 37.8417  \\
& Ours& \boldblue{50.5714} & \boldblue{44.3980} & \boldblue{50.6827} & \boldblue{42.1217}    
\\
\bottomrule
\end{tabular}
}

\vspace{-2mm}
\end{table}
\begin{figure}
    \centering
    \vspace{-2mm}
    \includegraphics[width=0.55\linewidth]{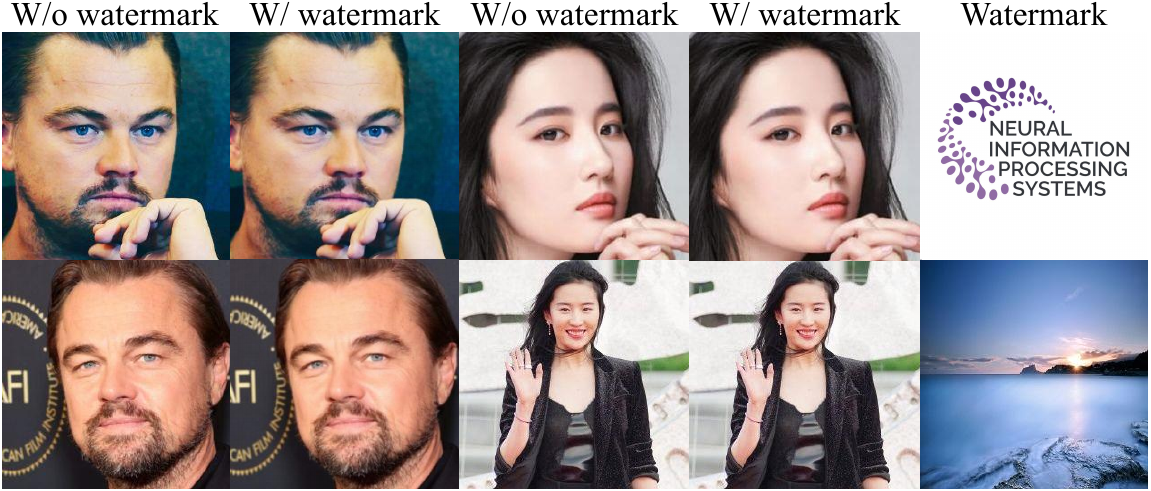}
    \caption{Visualized results of the original and watermarked portraits, the residual image of them, and the ground truth watermark \textit{NeurIPS} and \textit{IceShore}. "W/" and "W/o" denotes "with" and "without".}
    \label{tab:visual_watermark}
    \vspace{-5mm}
\end{figure}

\ding{113} \textbf{Dataset}
Due to the lack of existing data for IP protection, we collect a set of publicly available photos of celebrities from the Internet to serve as the IP information requiring protection, thereby proposing an IP protection dataset. Considering the diversity and fairness of the data, we select portraits of celebrities from different genders and skin colors, labeled as "Diva" (175 images), "Sportsman" (94 images), "Actor" (177 images), and "Actress" (206 images)\footnote{To protect the privacy of the celebrities, their real names are not used.}. All images are cropped to include the face and its surrounding area, and resized to 256$\times$256 pixels 
Then for each IP, we utilize four personalized generation models (LoRA~\cite{hu2022lora}, InstantID~\cite{wang2024instantid}, PhotoMaker~\cite{li2023photomaker}, DreamBooth~\cite{ruiz2023dreambooth}), with each method producing 1,000 images separately, and the prompts are generated by Large-Language-Models (LLMs) like ChatGPT~\cite{chatgpt}.
We split the dataset into training and validation sets in an 8:2 ratio.
For the images of copyright watermark, we use the logo of NeurIPS and a photo of ice shore, denoted as \textit{NeurIPS} and \textit{IceShore} in the following experiments. More details of our dataset are in Appendix~\ref{sec:appendix_dataset}.

\ding{113} \textbf{Network Structure}
Our network architecture consists of an embedding and decoder for the watermark, as well as a classification network for the specific attribution generation task. For the watermark network, we employ the reversible network structure from EditGuard~\cite{zhang2023editguard}. Regarding the attribution network, we follow the setting of~\cite{xu2024detecting} and utilize EfficientFormer~\cite{li2022efficientformer} as the backbone network for the vision task. All networks are trained from scratch. The detailed network architectures and training process of watermark embedding can be found in the Appendix~\ref{sec:network_architecture}.

\ding{113} \textbf{Settings}
For the watermark source-tracing task, we set the learning rate to 1e-4, batch size=1, and use three kinds of degradations, including Gaussian noise, Poisson noise, and JPEG compression, to simulate the loss of image quality in real network transmissions. We set the variation of Gaussian noise from 1 to 16, and JPEG compression quality randomly from 70 to 95. 
We set the learning rate to 2e-3 and batch size=32 for the attribution task of specific generation methods. 
In the incremental learning, we additionally generated 200 images for each method, with the remaining settings same as above.
All experiments are done on 4 NVIDIA 3090 GPUs.

\subsection{Evaluation of Source-Tracing and Attribution Task}
\paragraph{Evaluation of Source-Tracing}

\begin{table}
\footnotesize
    \centering
    \vspace{-1mm}
    \caption{Visualized results of source-tracing, on our proposed framework and Zhang et al.~\cite{9373945}. "Zhang et al.~\cite{9373945}$^{\dagger}$" indicates the box-free approach in Zhang et al.~\cite{9373945}, with our watermark encoder and decoder architecture. }
    \small
    \label{tab:visual_sta}
    \resizebox{0.75\linewidth}{!}{
    \begin{tabular}{c@{\hspace{0.1em}}*{7}{T}}
        Generated & Ours  & Zhang et al.~\cite{9373945} & Zhang et al.~\cite{9373945}$^{\dagger}$ &
         Generated & Ours  & Zhang et al.~\cite{9373945} & Zhang et al.~\cite{9373945}$^{\dagger}$ \\
        \includegraphics[width=0.139\textwidth]{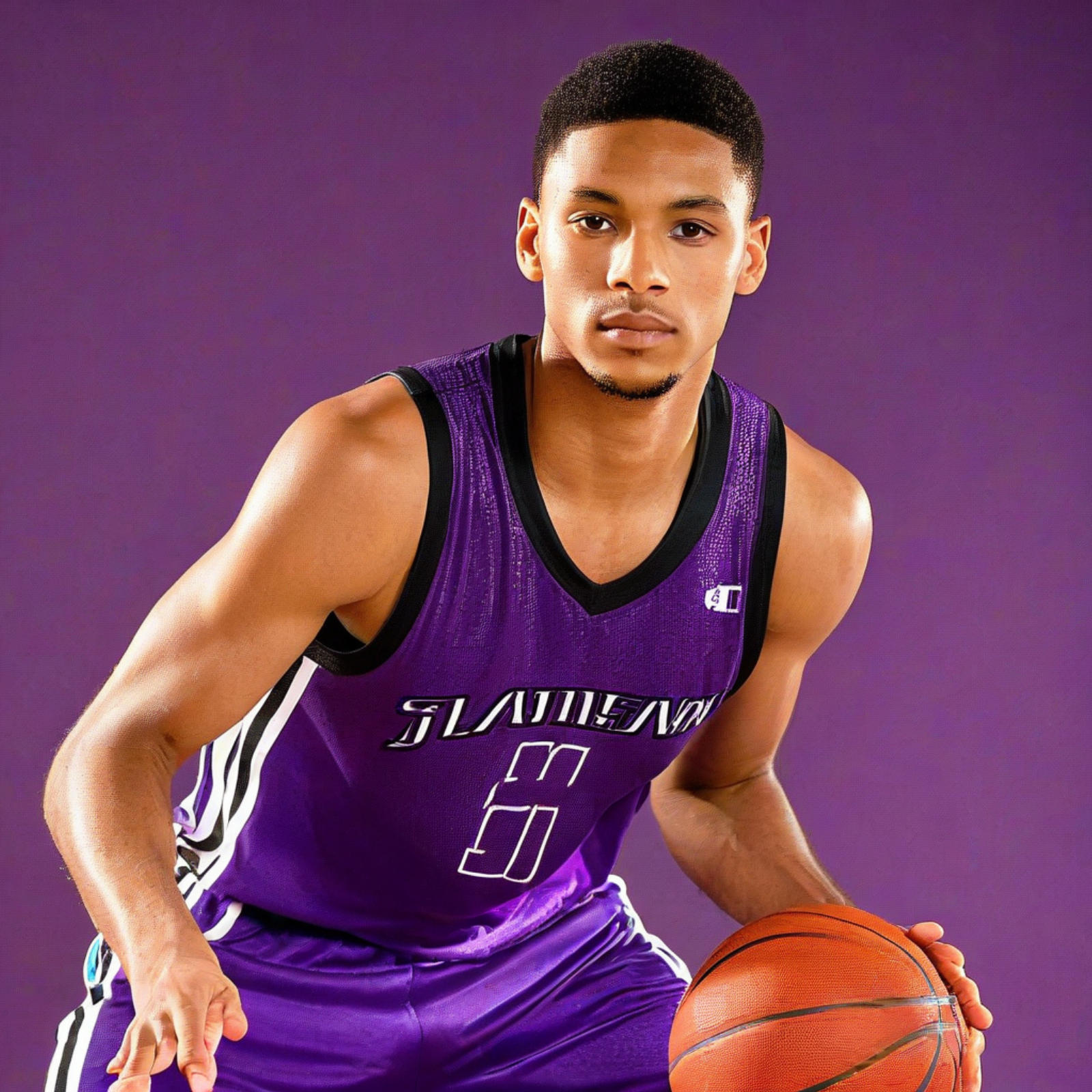} \hspace{-1mm} &
        \includegraphics[width=0.139\textwidth]{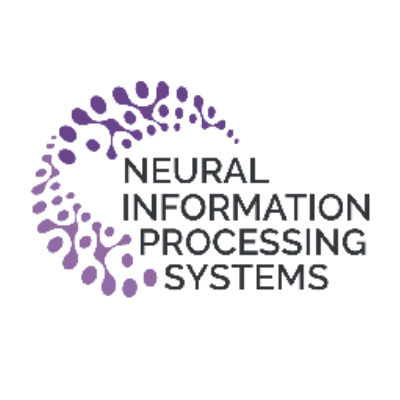} \hspace{-1mm} &
        \includegraphics[width=0.139\textwidth]{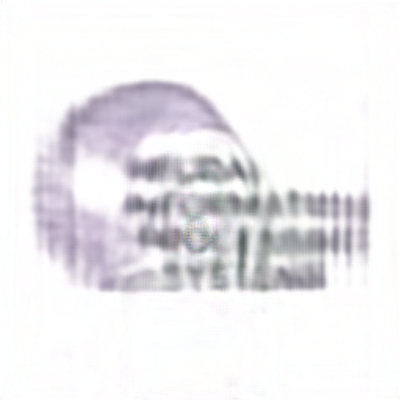} \hspace{-1mm} &
        \includegraphics[width=0.139\textwidth]{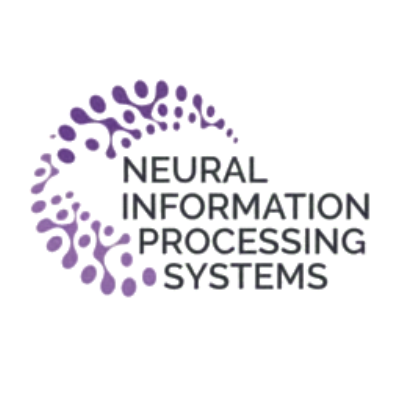} \hspace{-1mm} &
        \includegraphics[width=0.139\textwidth]{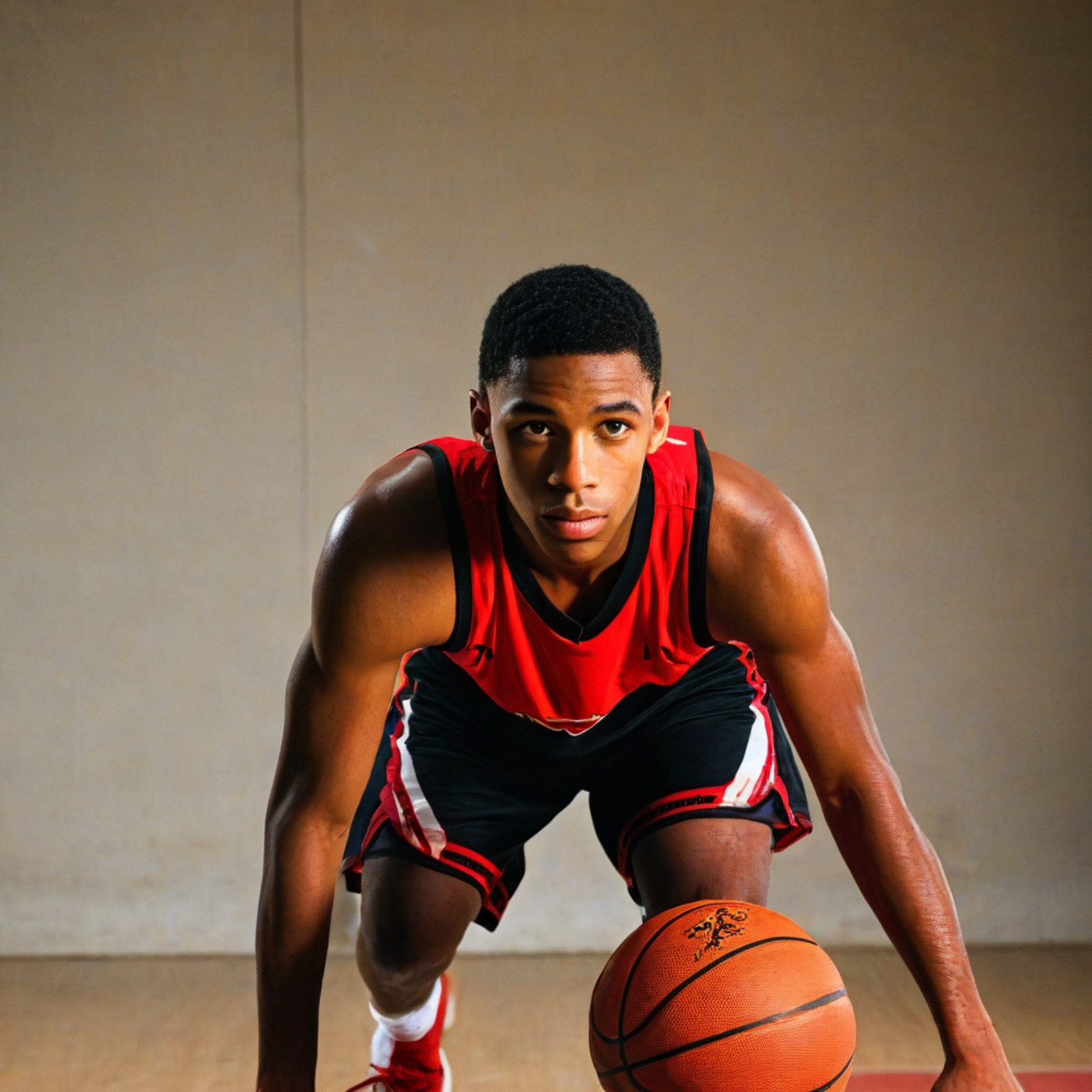} \hspace{-1mm} &
        \includegraphics[width=0.139\textwidth]{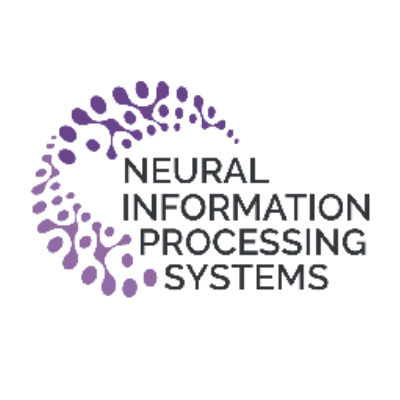} \hspace{-1mm} &
        \includegraphics[width=0.139\textwidth]{figure/zhang_get.pdf} \hspace{-1mm} &
        \includegraphics[width=0.139\textwidth]{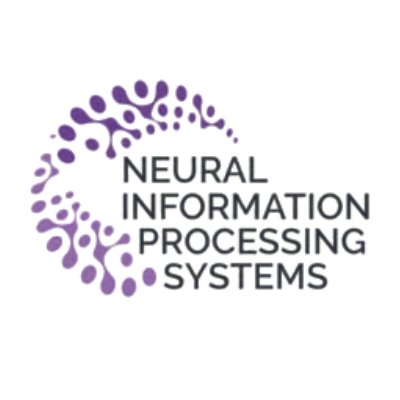} \hspace{-1mm} 
        \\
        \includegraphics[width=0.139\textwidth]{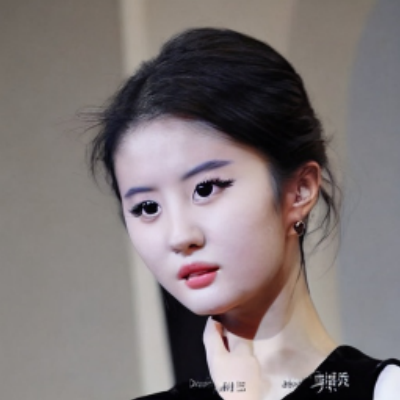} \hspace{-1mm} &
        \includegraphics[width=0.139\textwidth]{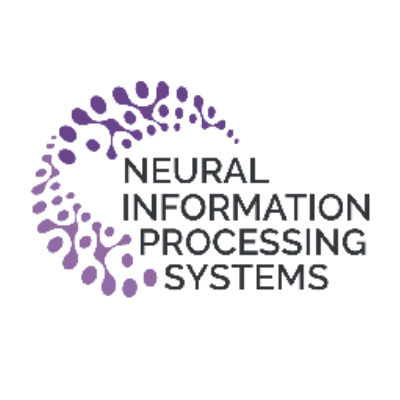} \hspace{-1mm} &
        \includegraphics[width=0.139\textwidth]{figure/zhang_get.pdf} 
        \hspace{-1mm} &
        \includegraphics[width=0.139\textwidth]{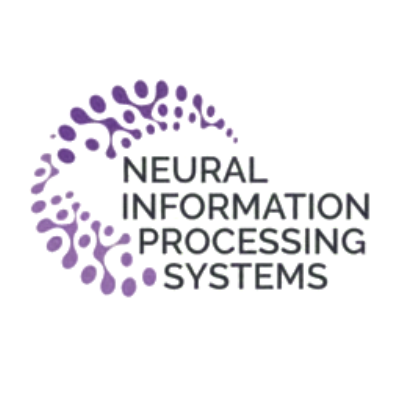} \hspace{-1mm} &
        \includegraphics[width=0.139\textwidth]{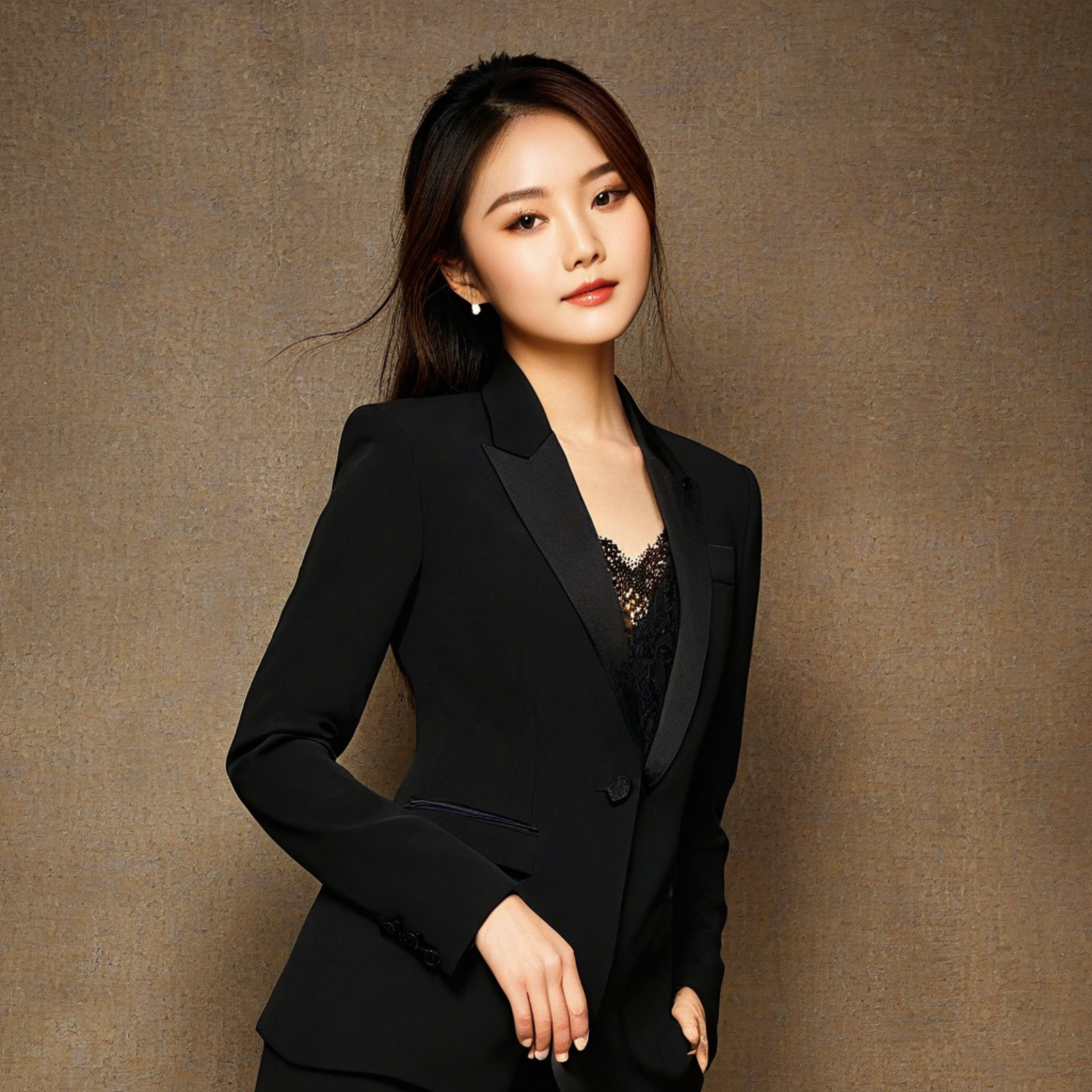} \hspace{-1mm} &
        \includegraphics[width=0.139\textwidth]{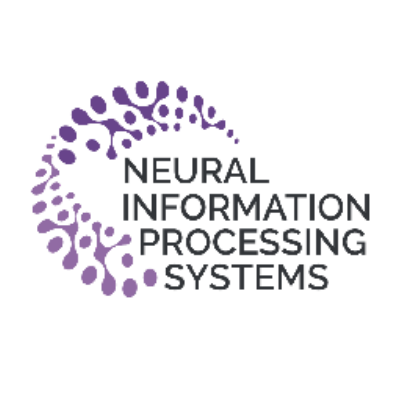} \hspace{-1mm} &
        \includegraphics[width=0.139\textwidth]{figure/zhang_get.pdf} \hspace{-1mm} &
        \includegraphics[width=0.139\textwidth]{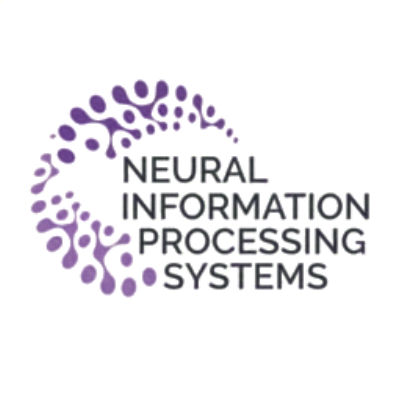} \hspace{-1mm} 
    \end{tabular}  
    }
    \vspace{-6mm}
\end{table}

\begin{table}
\small
\vspace{-5mm}
\caption{Evaluation of attribution task of \textit{passive} method, \textit{proactive} method, and our proposed method with unified \textit{passive} and \textit{proactive} mechanism. The metrics are attribution accuracy of (1) AIGC detection; (2) method judge, determining the specific generation method; and (3) overall attribution accuracy. Please note that the accuracy of method judge task is same for EfficientFormer~\cite{li2022efficientformer} and our method. "-" denotes results not available, as Prompt Inversion cannot achieve AIGC detection task. Best results are shown in \textbf{\textcolor{blue}{bold}}.}
\label{tab:attribution}
\centering
\resizebox{1\linewidth}{!}{
\begin{tabular}{cccccc}
\\
\toprule
Portrait IP & Method Type & Method & AIGC Detection & Method Judge & Overall Accuracy  \\
\midrule
\multirow{6}{*}{Diva}
&\multirow{4}{*}{Passive}
&ResNet~\cite{he2016resnet}&23.58\%&33.34\%&24.96\%\\
&&KNN~\cite{guo2003knn}&92.36\%&65.03\%&63.50\%\\
&&Prompt Inversion~\cite{li2024regeneration}&-&69.54\%&-  \\
&&EfficientFormer~\cite{li2022efficientformer} &86.28\%&\boldblue{95.00\%}& 91.80\%  \\
\cmidrule{2-3}
&Proactive
&Multi-Watermark&82.14\%&78.57\%& 80.01\%  \\
\cmidrule{2-3}
&Proactive \& Passive
&Ours&\boldblue{96.77\%}&\boldblue{95.00\%}& \boldblue{95.92\%} \\
\midrule
\multirow{6}{*}{Sportsman} 
&\multirow{4}{*}{Passive}
&ResNet~\cite{he2016resnet}  &24.61\%&34.97\%& 25.00\%   \\
&&KNN~\cite{guo2003knn} &97.27\%&58.40\%& 59.04\%    \\
&&Prompt Inversion~\cite{li2024regeneration} &-&61.33\%& -    \\
&&EfficientFormer~\cite{li2022efficientformer} &90.96\%&\boldblue{95.80\%}& 90.97\% \\
\cmidrule{2-3}
&\multirow{1}{*}{Proactive}
&Multi-Watermark&89.58\%&51.76\%& 59.76\%  \\
\cmidrule{2-3}
&\multirow{1}{*}{Proactive \& Passive}
&Ours &\boldblue{98.05\%}&\boldblue{95.80\%}&\boldblue{96.69\%}   \\
\midrule
\multirow{6}{*}{Actor}  
&\multirow{4}{*}{Passive}
&ResNet~\cite{he2016resnet} &24.92\%&31.10\%& 24.93\%  \\
&&KNN~\cite{guo2003knn}  &89.45\%&67.86\%& 68.40\%\\
&&Prompt Inversion~\cite{li2024regeneration}  &-&58.66\%& - \\
&&EfficientFormer~\cite{li2022efficientformer} &87.03\%&\boldblue{98.83\%}& 87.10\% \\
\cmidrule{2-3}
&\multirow{1}{*}{Proactive}
&Multi-Watermark&86.88\%&50.81\%& 59.43\%  \\
\cmidrule{2-3}
&\multirow{1}{*}{Proactive \& Passive}
&Ours&\boldblue{93.54\%}&\boldblue{98.83\%}& \boldblue{96.07\%}   \\
\midrule
\multirow{6}{*}{Actress}  
&\multirow{4}{*}{Passive}
&ResNet~\cite{he2016resnet} &23.67\%&34.98\%& 24.90\%   \\
&&KNN~\cite{guo2003knn}  &92.17\%&66.63\%& 67.46\% \\
&&Prompt Inversion~\cite{li2024regeneration}  &-&62.00\%& - \\
&&EfficientFormer~\cite{li2022efficientformer} &87.09\%&\boldblue{93.38\%}&   89.69\%  \\
\cmidrule{2-3}
&\multirow{1}{*}{Proactive}
&Multi-Watermark&92.06\%&65.08\%&76.06\%  \\
\cmidrule{2-3}
&\multirow{1}{*}{Proactive \& Passive}
&Ours&\boldblue{92.55\%}&\boldblue{93.38\%}& \boldblue{92.99\%}\\
\bottomrule
\end{tabular}}
\vspace{-3mm}
\end{table}


For the box-free watermark source-tracing method, we compared against the approach of Zhang et al.~\cite{9373945}, with evaluation metrics including (1) the difference between the image before and after watermark embedding, and (2) the accuracy of decoding the watermark from generated images $\mathbf{I}_G$. The quantitative results are presented in Tab.~\ref{tab:main}, and the visualization of watermarking embedding and source-tracing is demonstrated in Fig.~\ref{tab:visual_watermark} and Tab.~\ref{tab:visual_sta}. Both quantitative and qualitative results have shown that our watermark framework can preserve the quality of images, and can be extracted correctly. We have also done a security analysis for our watermarking method and Zhang et al.~\cite{9373945}, and detailed description and results are in Appendix~\ref{sec:appendix_security}.

\paragraph{Evaluation of Attribution}
\label{sec:evaluation_attribution}
For the attribution task, we compared common visual baselines, including the supervised ResNet~\cite{he2016resnet} and the clustering method KNN~\cite{guo2003knn}, to demonstrate that this task is challenging and non-trivial. We also compare our method to Prompt Inversion~\cite{li2024regeneration}, which uses LLM to obtain possible prompts for the AIGC image and use it to find the most similar generative model. Subsequently, to validate the effectiveness of our proposed proactive watermarking and passive detection methods, we compared them with the direct use of EfficientFormer for 4-class classification (1 label for true image and 3 labels for 3 generation models). Additionally, a trivial approach is to directly let the decoder output different images for different generation methods, thereby simultaneously achieving copyright source-tracing and method attribution, named as \textit{Multi-Watermark}. While this is indeed a clever approach, we compare this potential approach and find that it could not effectively accomplish the task. The results of the comparison task are shown in Tab.~\ref{tab:attribution}.

Please note that there are some \textit{box-free watermarking}, \textit{attribution}, and \textit{source-tracing} methods that are not included in our evaluation results, and the comparison and explanation of why they are not available in our settings are listed in Appendix~\ref{sec:appendix_notcompare}.


\begin{wraptable}{r}{7cm}
\vspace{-5mm}
\caption{Effect of incremental learning strategy. Metric \textit{Knowledge Forget} indicates performance reduction during fine-tuning. Best results are shown in \textbf{\textcolor{blue}{bold}}.}
\small

\label{tab:incremental}
\centering
\newcommand{\tabincell}[2]{\begin{tabular}{@{}#1@{}}#2\end{tabular}}
\resizebox{1.\linewidth}{!}
{
\begin{tabular}{ccc}
\toprule
Portrait IP     & Method & Knowledge Forget $\downarrow$  \\
\midrule
\multirow{2}{*}{Diva} 
&   Vanilla Fine-Tuning     & 4.11\%       \\
&   Incremental Strategy (Ours)     & \boldblue{0.35\%}         \\
\midrule
\multirow{2}{*}{Sportsman} 
&   Vanilla Fine-Tuning     & 25.63\%                \\
&   Incremental Strategy (Ours)     & \boldblue{0.71\%}         \\
\midrule
\multirow{2}{*}{Actor} 
&   Vanilla Fine-Tuning     & 3.39\%                \\
&   Incremental Strategy (Ours)     & \boldblue{1.52\%}         \\
\midrule
\multirow{2}{*}{Actress} 
&   Vanilla Fine-Tuning     & 8.27\%     \\
&   Incremental Strategy (Ours)   & \boldblue{2.29\%}     \\
\bottomrule
\end{tabular}
}

\vspace{-5mm}
 \end{wraptable}

\subsection{Ablation Studies}
In this subsection, we demonstrate the enhancement achieved through the joint proactive watermarking and passive detection mechanism, as well as the reduction in knowledge forgetting brought about by incremental learning strategy compared to direct fine-tuning methods.


\ding{113} \textbf{Effect of Incremental Learning Strategy}

For the attribution of newly appeared personalized models, we fine-tune the models obtained in Sec.~\ref{sec:attribution} using (1) direct fine-tuning and (2) incremental learning strategies separately. The quantitative results are shown in Tab.~\ref{tab:incremental}, which shows that our incremental strategy helps the attribution network mitigate the knowledge forgetting significantly. The metric \textit{Knowledge Forget} is calculated by $(1-a_{new}/a_{original})$, where $a_{new}$ is the accuracy of the fine-tuned model, and $a_{original}$ for original model~\cite{he2020incremental}.

\ding{113} \textbf{Choice of Hyper-parameter}
\label{sec:lambda_c}
In Sec.~\ref{sec:exp_increment}, we introduced the loss function to fine-tune the original attribution network with an extra supplement dataset, to reduce the knowledge forgetting. This loss function incorporates a hyper-parameter denoted as  
$\lambda_c$, balancing the two constituent parts of the loss. To validate the influence of $\lambda_c$, we perform a grid-search on it using "Sportsman" IP, with the findings depicted in Fig.~\ref{fig:ablation_lambda_c}.

\subsection{Method Analysis}
\label{sec:method_analysis}
\begin{figure}[htbp]
\centering
\vspace{-3mm}
\begin{minipage}[t]{0.48\textwidth}
\centering
\includegraphics[width=6.5cm]{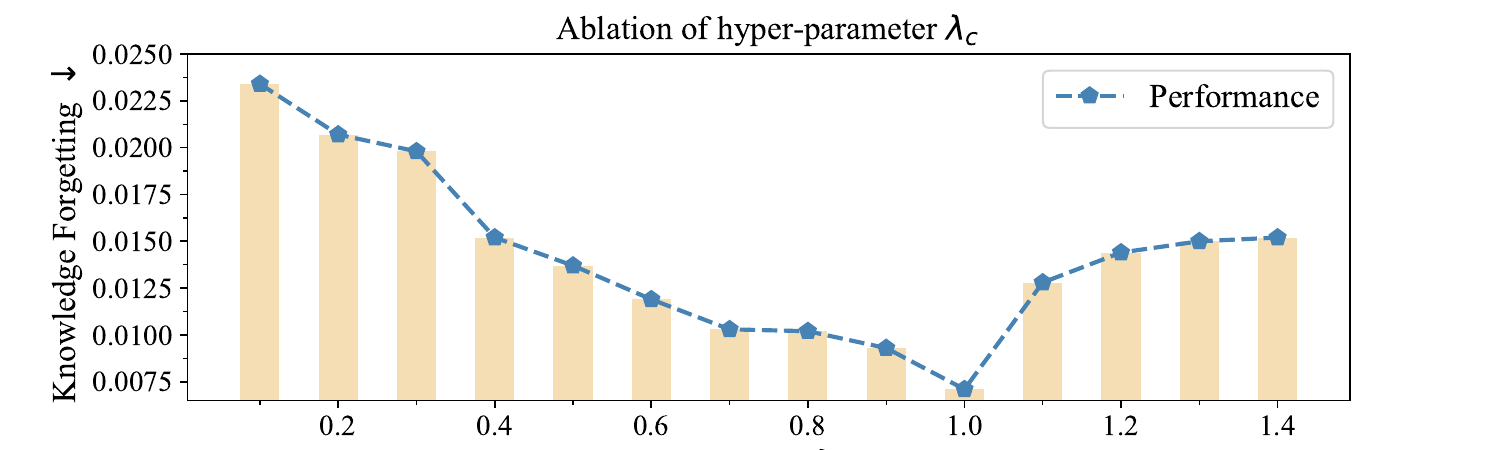}
\caption{Visualized statistics of different $\lambda_c$ on incremental learning strategy. It is shown that when $\lambda_c=1.0$, the knowledge forgetting caused by fine-tuning is minimal.}
\label{fig:ablation_lambda_c}
\end{minipage}
\hspace{1mm}
\begin{minipage}[t]{0.48\textwidth}
\centering
\includegraphics[width=6.5cm]{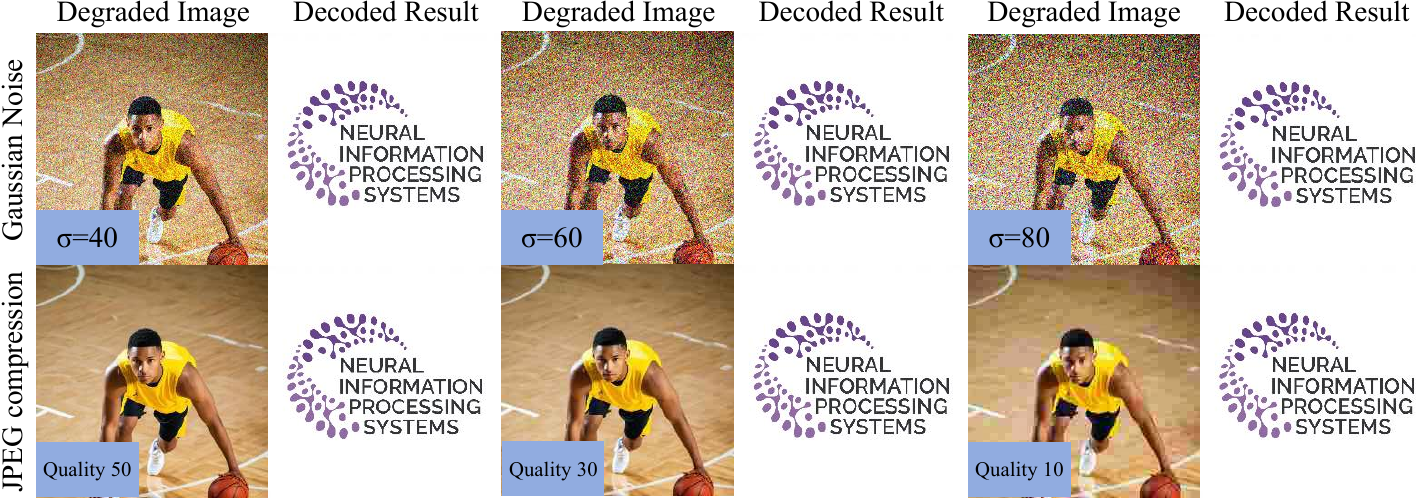}
\caption{Result of robustness analysis on our proposed framework. We add severe degradations separately on generated images, and test whether our watermark is extracted correctly.}
\label{fig:robust}
\end{minipage}
\hspace{.15in}

\vspace{-8mm}
\end{figure}

\ding{113} \textbf{Robustness Analysis} To evaluate the robustness of our source-tracing watermark, we have subjected the images in the test set to Gaussian noise and JPEG compression, with degradation intensities exceeding those set during the training process. The visualization of the effects is depicted in the accompanying Fig.~\ref{fig:robust}. Results show that our watermark is enough robust to common degradations on Internet transmission.

\ding{113} \textbf{Different Base Models in Personalized Generation} 
Some newly personalized generative models, such as PhotoMaker~\cite{li2023photomaker}, use pre-trained base models and generate images via the base model. Assuming that the stealer trains a base model on their own, instead of the one provided by the official code or API, it could lead to failure in source-tracing. We test this possible scenario using PhotoMaker as an example, and detailed results are in Appendix~\ref{sec:appendix_base}, which shows that our watermark is detectable to unseen base models.

\section{Conclusion}
To safeguard the intellectual property (IP) and portrait rights of images, we introduce the task of copyright source-tracing and attribution of AIGC misuse methods in this paper. We propose a novel watermarking method that combines proactive and passive approaches, enabling the decoding of copyright watermarks from images generated by personalized models and identifying the specific generation methods. Furthermore, we employ an incremental learning strategy to efficiently attribute newly emerging personalized methods, thereby avoiding catastrophic forgetting issues caused by new data. To evaluate our proposed method, we prepare a dataset consisting of a series of portraits and their corresponding personalized generation results. Experimental results on this dataset demonstrate the effectiveness of our proposed approach. Our method can be extended to various applications involving copyright protection and defense against illegal personalized model generation.

\bibliographystyle{plain}
\bibliography{example.bib}
\appendix
\section*{Appendix / supplemental material}
In this Appendix part, we will supplement contents including (1) additional experiments about our method in Appendix~\ref{sec:appendix_additionalexp}; (2) the comparison of our method and other related work that are similar but not applicable in comparison experiments in Appendix~\ref{sec:appendix_notcompare}; (3) introduction of personalized generation methods in Appendix~\ref{sec:appendix_p}; (4) detailed network architecture and training of watermark embedding in Appendix~\ref{sec:network_architecture}; (5) Details of collecting and personalized generation of our IP dataset in Appendix~\ref{sec:appendix_dataset}; (6) limitations and discussion of our work in Appendix~\ref{sec:limitation}.
\section{Additional Experiment Results}
\label{sec:appendix_additionalexp}
\begin{figure}
    \centering
    \includegraphics[width=1\linewidth]{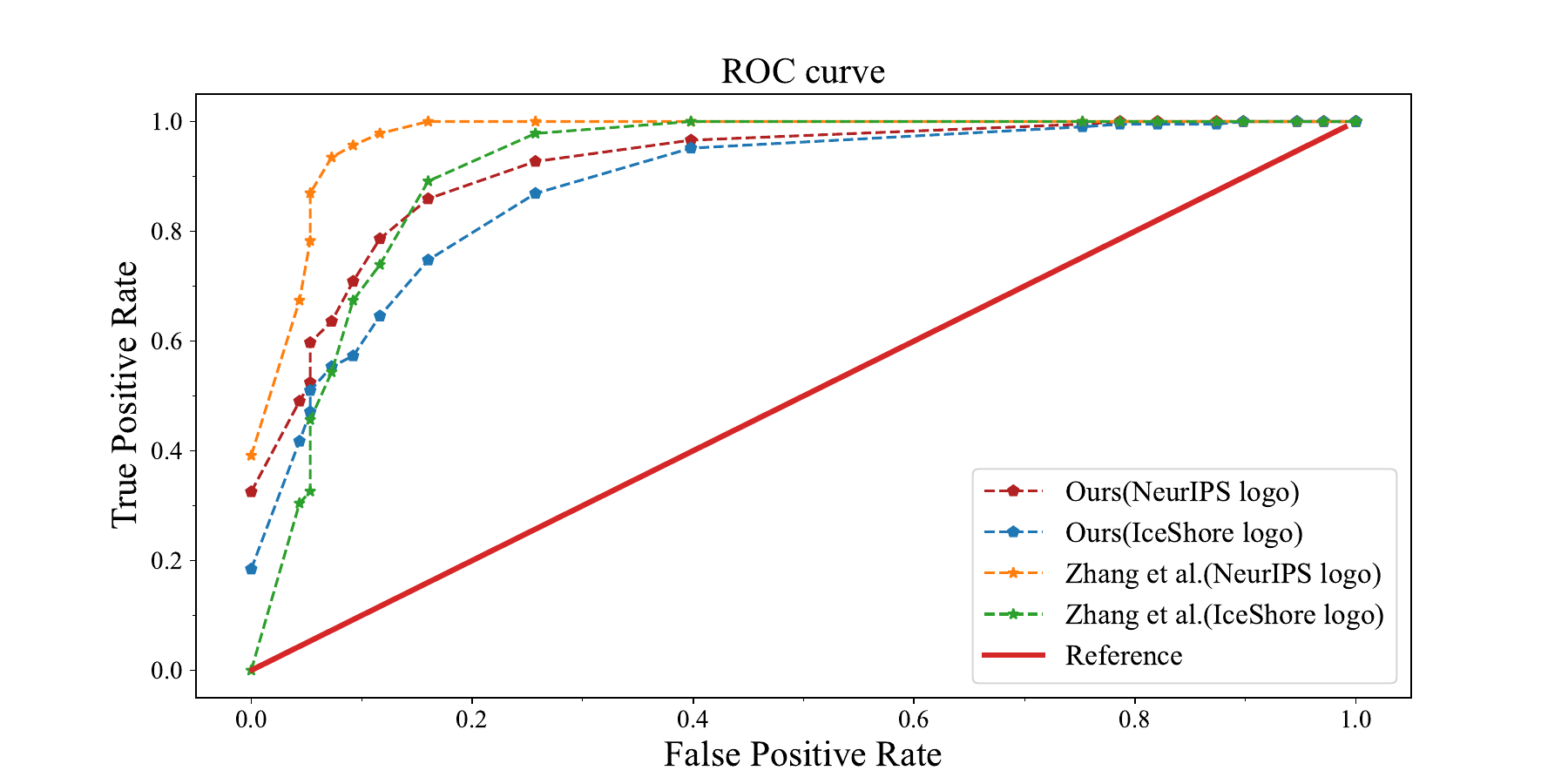}
    \caption{Security analysis of our proposed watermark.}
    \label{fig:security}
\end{figure}

\paragraph{Analysis of Different Base Models}
\label{sec:appendix_base}
We take PhotoMaker as an example, selecting YamerMIX\footnote{Link of YamerMIX on CivitAI is \href{https://civitai.com/models/84040/sdxl-unstable-diffusers-yamermix}{here}.} as the base model for testing, which is different from RealVisXL\footnote{Link of RealVisXL on HuggingFace is \href{https://huggingface.co/SG161222/RealVisXL_V3.0}{here}.} used during training. There are noticeable differences in the background and style of the generated results between these two models, with specific visualizations shown in the Tab.~\ref{tab:visual_base}. From the results, it can be seen that our method can also successfully trace the source for unseen base models.
\begin{table}
\footnotesize
    \centering
    \caption{Visualized results of source-tracing with used base model unseen in the training process.}
    \small
    \label{tab:visual_base}
    \begin{tabular}{c@{\hspace{0.1em}}*{3}{T}}
        \scriptsize{Generated Result} & \scriptsize{Decoded Watermark}  & \scriptsize{Generated Result} &
         \scriptsize{Decoded Watermark} \\
        \includegraphics[width=0.139\textwidth]{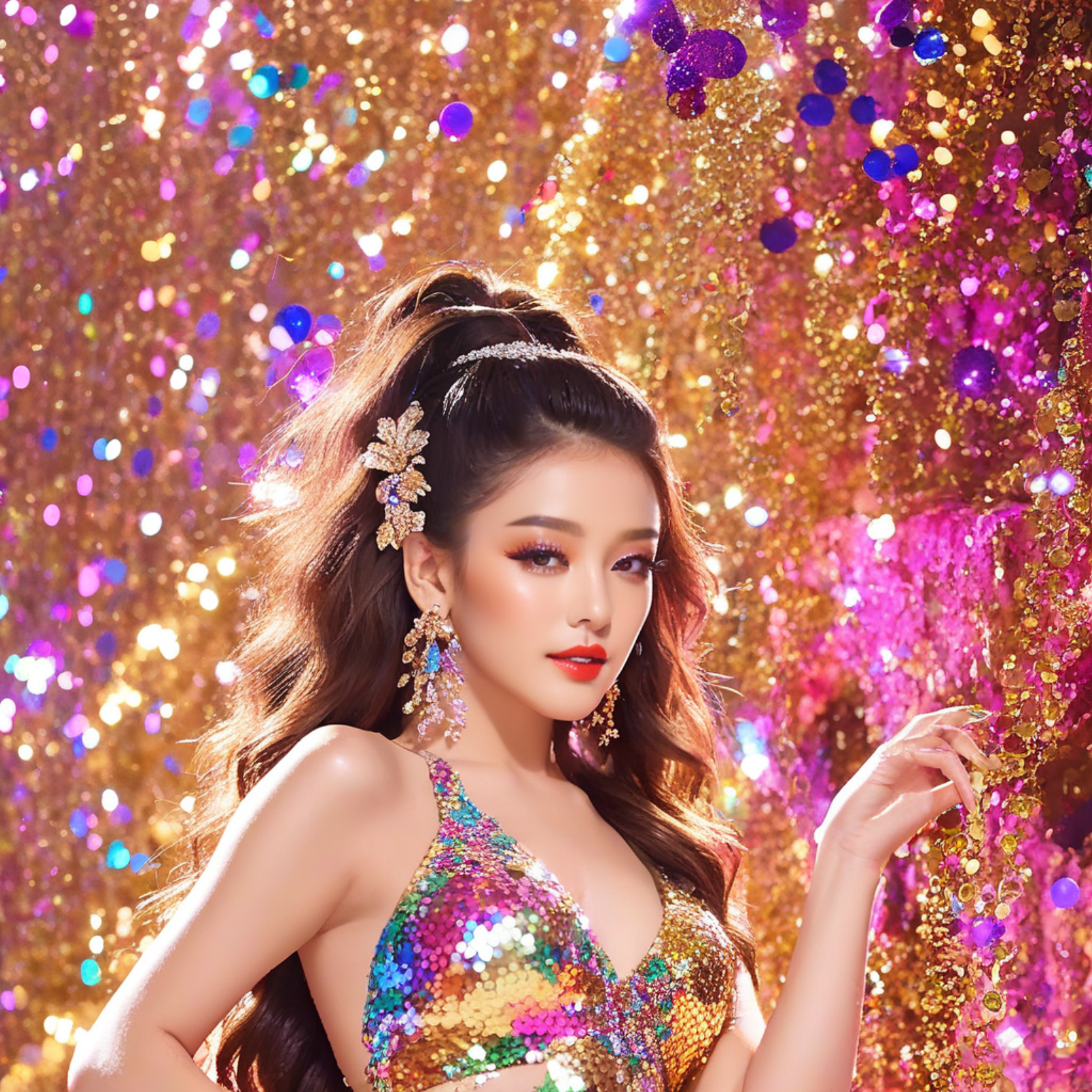} \hspace{-1mm} & 
        \includegraphics[width=0.139\textwidth]{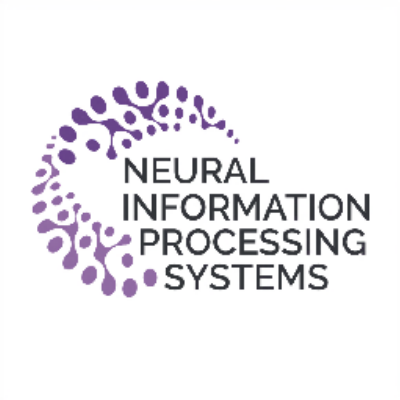} \hspace{-1mm} & 
        \includegraphics[width=0.139\textwidth]{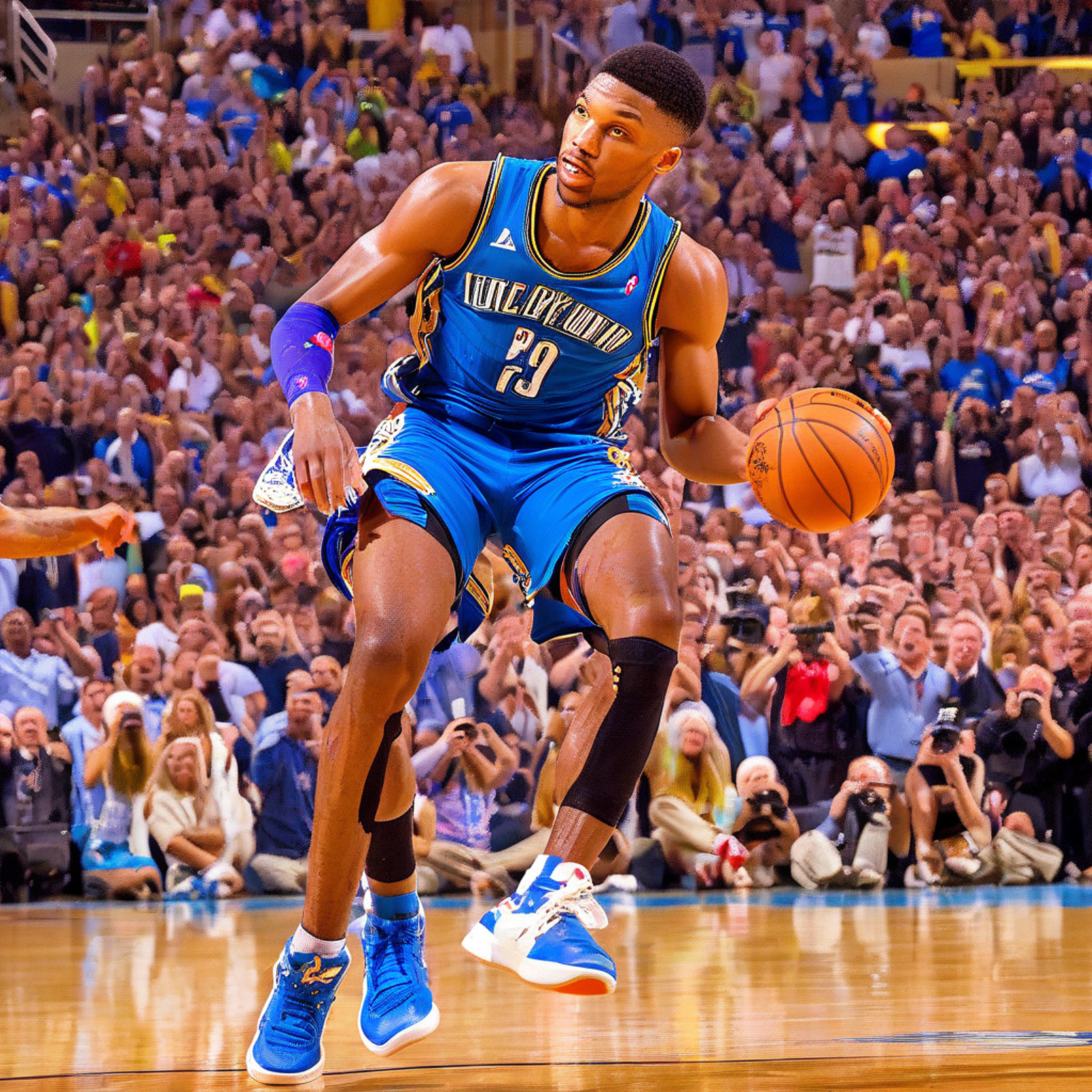} \hspace{-1mm} &
        \includegraphics[width=0.139\textwidth]{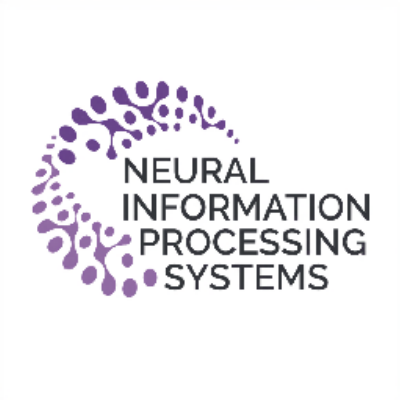} \hspace{-1mm} 
        \\
        \includegraphics[width=0.139\textwidth]{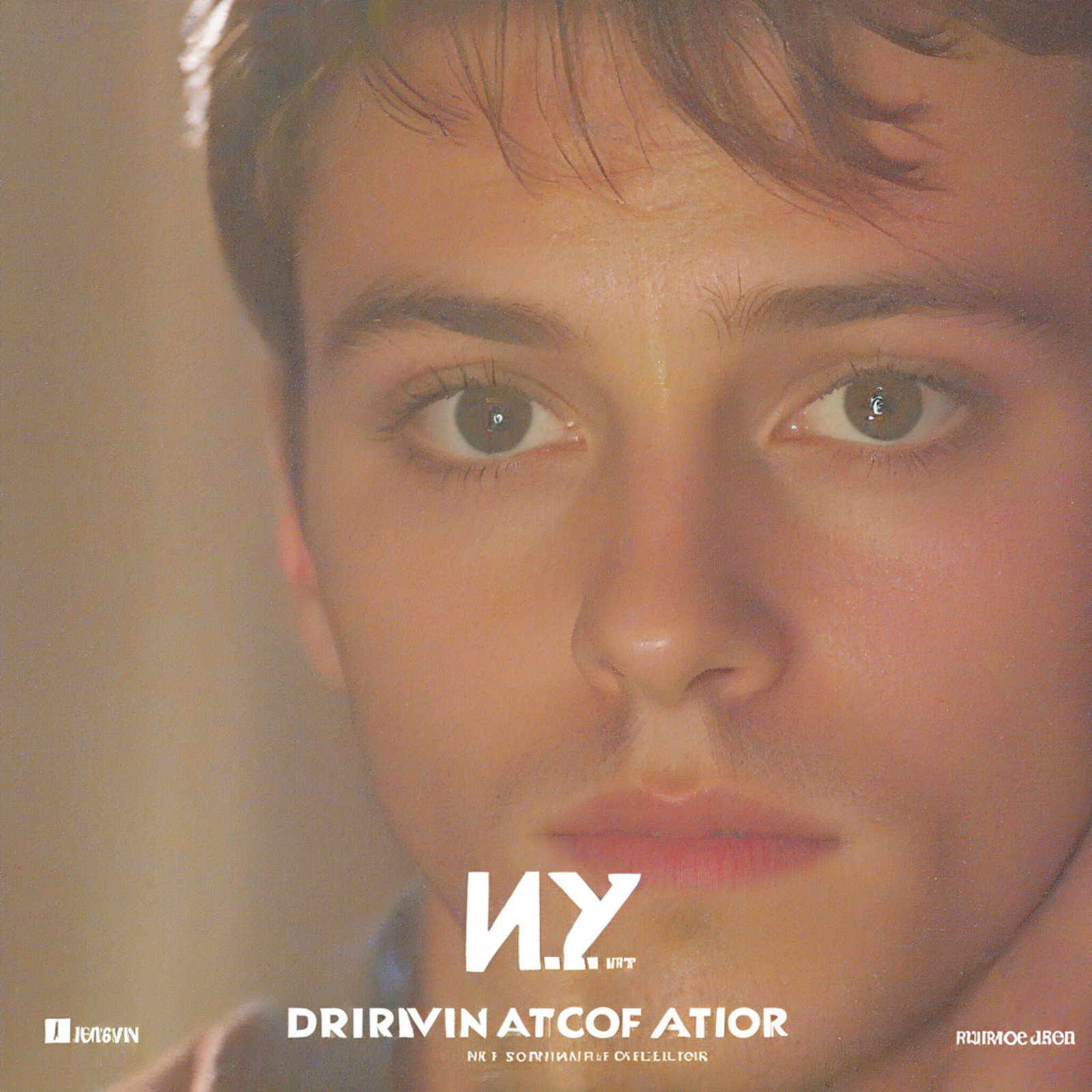} \hspace{-1mm} & 
        \includegraphics[width=0.139\textwidth]{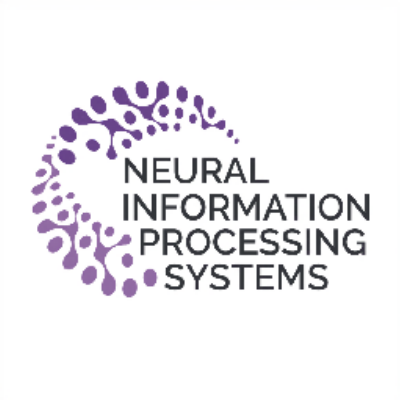} \hspace{-1mm}  &
        \includegraphics[width=0.139\textwidth]{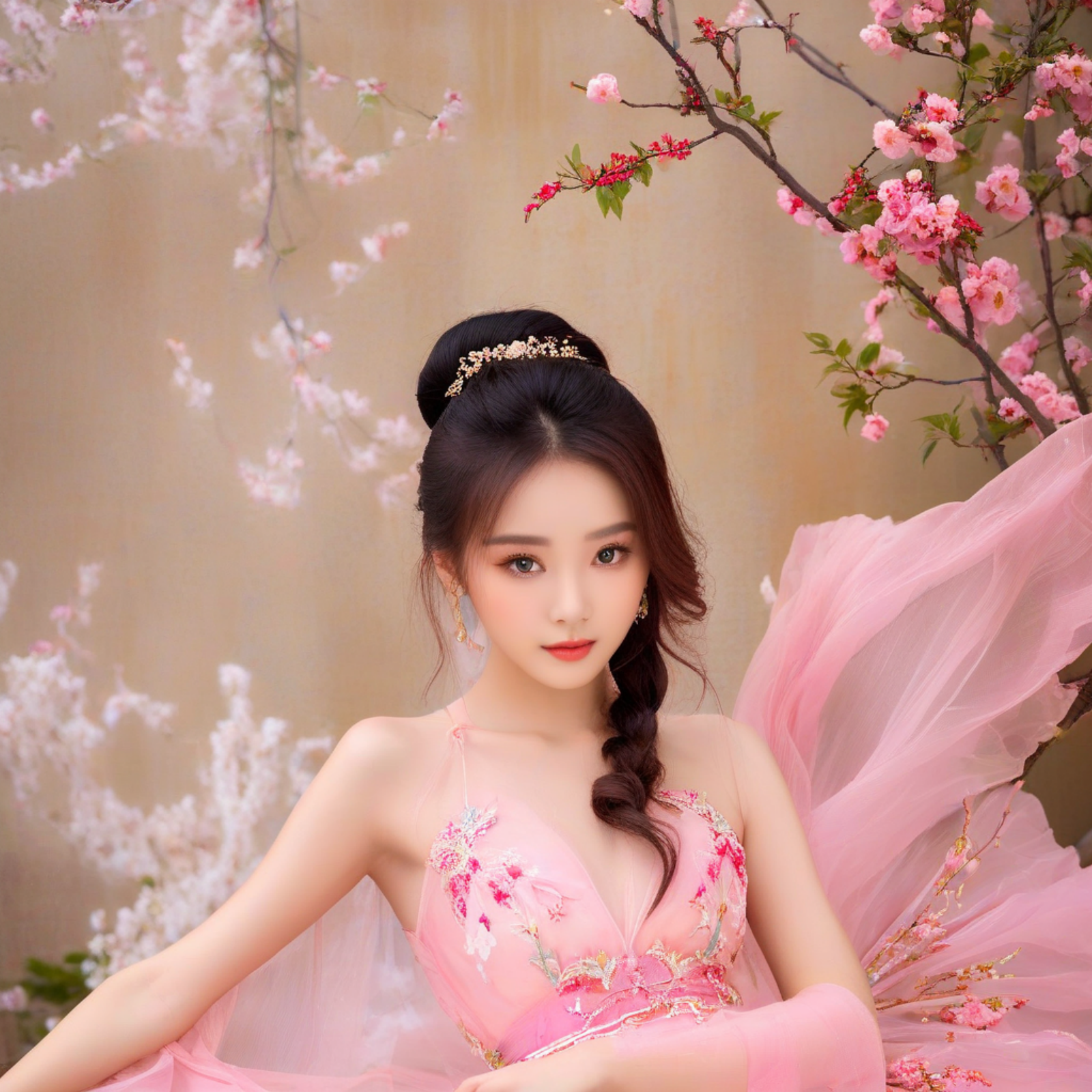} \hspace{-1mm} & 
        \includegraphics[width=0.139\textwidth]{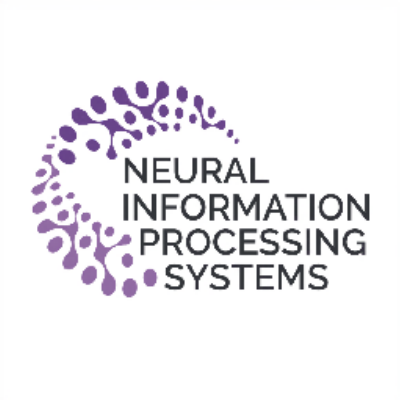} \hspace{-1mm}
        
        \\ 

    \end{tabular}  
    
    \vspace{-3mm}
\end{table}
\paragraph{Security Analysis}
\label{sec:appendix_security}
To validate the security of our proposed watermark framework, we perform an anti-steganography detection using StegExpose~\cite{boehm2014stegexpose} on watermarked images of Zhang et al.~\cite{9373945} and our framework. Both of them embed a 256$\times$256 RGB image. The detection set is built by mixing watermarked images and original images with equal proportions. We vary the detection thresholds in a wide range in StegExpose~\cite{boehm2014stegexpose} and draw the receiver operating characteristic (ROC) curve in Fig.~\ref{fig:security}. Please note that the closer the curve in the figure is to the \textbf{reference} line, the more difficult it is for the detecting model to detect the method corresponding to that curve, and thus the watermarking method is considered to be more secure. 
(The ideal case for a watermarking method is nearly close to the reference line, meaning a 50\% accuracy of judging whether the image has embedded a watermark, which is random-guess). Fig.~\ref{fig:security} shows that for watermark \textit{NeurIPS} and \textit{IceShore}, the curve of our method is closer to the reference line than Zhang et al.~\cite{9373945}, demonstrating the reliable security of our proposed watermark.

\section{Explanation of Methods Not Available in Experiments}
\label{sec:appendix_notcompare}
For the box-free watermarking method Huang et al.~\cite{huang2023can}, their work is focused on copyright verification of GAN-based methods, and a specific discriminator for GAN architecture is needed, while our work is focused on diffusion-based generation methods.
For attribution methods~\cite{yu2020responsible,yu2021artificial}, their attribution approach is based on GAN finger-printing, while our scenario is diffusion-based personalized generation methods; the watermark added in these works are bit messages, while our watermark is an image.
For source-tracing methods~\cite{wang2024must,wu2023sepmark}, the watermark added in these works are also bit messages, while our watermark is an image.
\section{Personalized Generation Methods}
\label{sec:appendix_p}
Diffusion Probabilistic models~\cite{ho2020denoising, song2021denoising,rombach2021highresolution} have introduced a new paradigm to generative models, particularly for personalized generative models~\cite{zhang2024surveypersonalized}. LoRA~\cite{hu2022lora} incorporates a low-rank representation into the diffusion process, enabling the generation of specific character traits with a relatively minor training cost. However, LoRA still necessitates training, and a high-performing LoRA model requires dozens to hundreds of images, along with several hours of training time. Additionally, it is necessary to prepare caption texts corresponding to these images.
DreamBooth\cite{ruiz2023dreambooth} learns the character information corresponding to a specific token, such as "sks person". By adding such token to the generation prompt in Stable Diffusion~\cite{rombach2021highresolution}, an image of that character can be generated. To reduce training costs and enhance the efficiency of generating personalized images, several training-free methods have been proposed.
InstantID~\cite{wang2024instantid} leverages a pre-trained ControlNet~\cite{zhang2023adding} model as its foundation, and with just a single specific image, it can generate a personalized image corresponding to that person's face ID. PhotoMaker~\cite{li2023photomaker}, on the other hand, can directly extract the character's face ID information from an image and generate a personalized image for that ID, allowing for the overlay of faces.

\section{Network Architecture and Training Watermark Embedding}
\label{sec:network_architecture}
Our watermark embedding and decoding network is referred from EditGuard~\cite{zhang2023editguard}. As we do not include bit message into the watermark, we modify the network into an invertible structure with \textit{images} as input and output, and the detailed architecture and training process are shown below:

\textbf{Watermark Encoder} It embeds a 2D image watermark original image, forming a container image. We do not use the bit encoder structure.

\textbf{Copyright Extractor} It extracts copyright information from the container image, which is robust against degradation including noise and JPEG compression.

\textbf{Invertible Blocks} They are used in the encoder and extractor for precise multimedia information recovery through Discrete Wavelet Transform (DWT) and enhanced affine coupling layers.

\textbf{Training Watermark Embedding}
The watermark encoder $\mathcal{E}$ takes as inputs an original image $\mathbf{I}_O$ and a watermark image $\mathbf{I}_W$. As we choose the invertible network as watermark encoder as well as decoder, by reversing the direction of input and output, we can decode the watermark image using the same network and shared weights. The training target consists of two parts: (1) decode the image correctly, and (2) the image embedded with the watermark $\mathbf{I}_{O}^{'}$ is similar to the original image. The specific loss is as Eq.~\ref{eq:embedding}:
\begin{equation}
\label{eq:embedding}
    \mathcal{L}_{\text{embedding}} = ||\mathbf{I}_O-\mathbf{I}_{O}^{'}||_2^2 + ||\mathcal{E}(\mathbf{I}_{O}^{'}),\mathbf{I}_W||_2^2
\end{equation}

\begin{figure}
    \centering
    \includegraphics[width=1\linewidth]{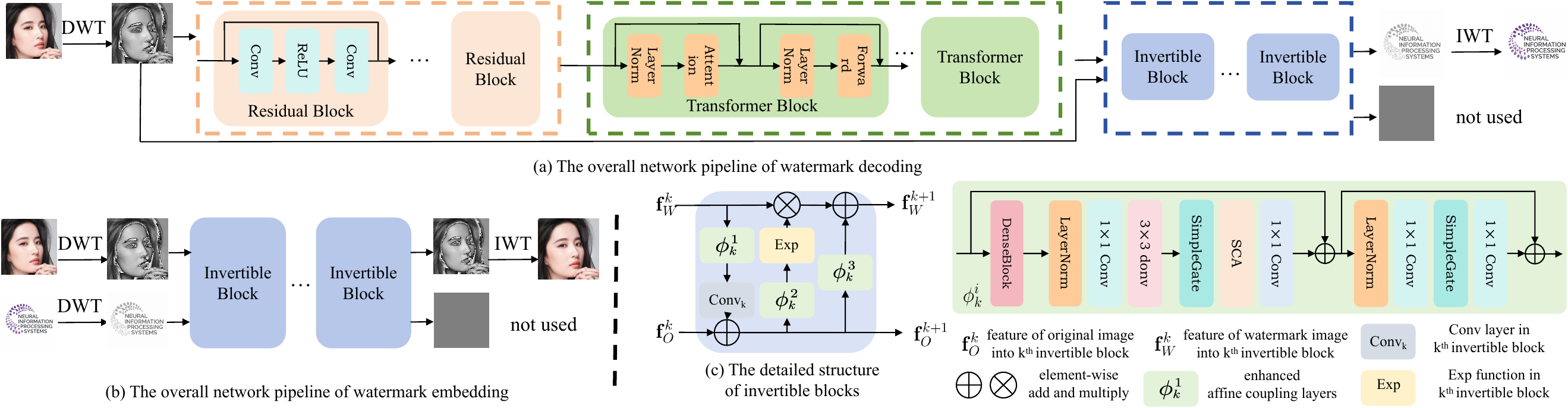}
    \caption{Detailed network structure of (a) watermark decoding pipeline; (b) watermark embedding pipeline; (c) structure of invertible blocks.}
    \label{fig:net}
\end{figure}

The network is designed to achieve precise copyright information recovery. It provides a proactive approach to revealing copyright information suitable for various AI-generated content (AIGC) methods. More detailed information is shown in Fig.~\ref{fig:net}.

\section{Details of Dataset Collection and Generation}
\label{sec:appendix_dataset}
We download photos of four well-known figures from websites such as Google Images, which are referred to as "Diva", "Sportsman," "Actor," and "Actress." For each downloaded image, we first filter out photos that are not suitable for training data, such as those with group photos or are blurry. Then, using OpenCV's face recognition method, we identify the faces in the images and crop them centered on the faces, ensuring that the resulting images are square images centered on the face, and finally resize them to 256*256. Each figure ended up retaining 175, 94, 177, and 206 photos respectively.

For the training of LoRA and DreamBooth, we use BLIP~\cite{li2022blip} image caption method to generate corresponding captions for each image. These image-text pairs are used for the training of LoRA and DreamBooth. PhotoMaker and InstantID do not require training.

\begin{figure}
    \centering
    \includegraphics[width=1\linewidth]{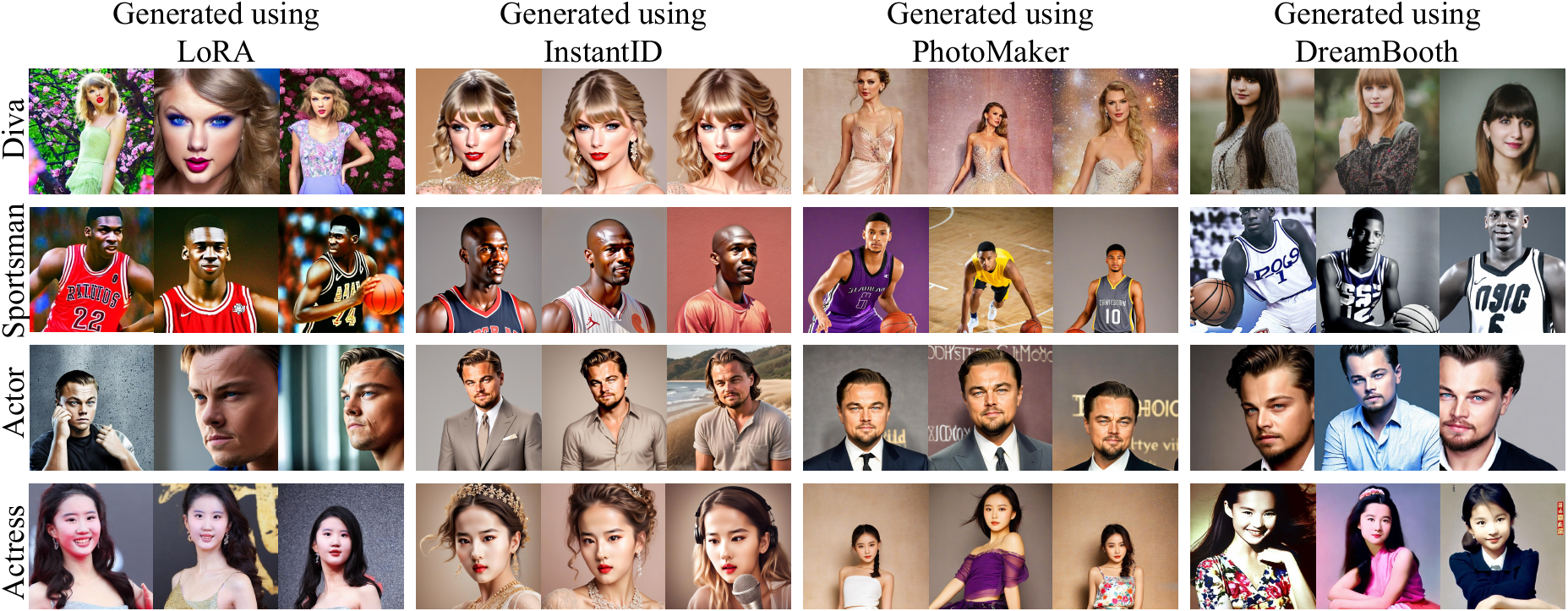}
    \caption{Images from four IPs and four personalized generation methods.}
    \label{fig:dataset}
\end{figure}

For the prompts used in image generation, we use Large-Language-Models (LLMs) like ChatGPT~\cite{chatgpt} to generate them, with specific prompts such as: "\textit{Please generate 100 prompts to generate a photo of an actor.}" Then, during actual generation, we add the following positive prompts: \textit{"high quality, high resolution, high definition, great face, colorful"}, and the following negative prompts: \textit{(asymmetry, worst quality, low quality, illustration, 3d, 2d, painting, cartoons, sketch), open mouth, grayscale}.
After all images are generated, they are randomly divided into training and validation sets in an 8:2 ratio. 
We select several images from the dataset and presented them in Fig.~\ref{fig:dataset}.

\section{Limitations and Discussion}
\label{sec:limitation}
Although our proposed method shows promising value in application of IP protection, there still exists potential limitations: (1) The dataset for training and validation is not extensive enough; (2) Our approach only supports method-specific detection and attribution, which needs an extra training process for new stealing methods, and it would be more flexible if a self-adaptive approach for most existing personalized generative methods is proposed. 

This study pioneers a method that combines proactive watermark detection with passive attribution to safeguard image IPs from illegal personalized generation. In future potential endeavors, our work could extend to other types and even other modalities of IP protection tasks. Furthermore, there is a lack of research on methods similar to GAN fingerprinting for diffusion model-based personalized generation models, which could differentiate between different models from the perspective of features in the model-generated results. This could enable training-free detection and attribution of AIGC models.


\end{document}